\documentclass[journal]{IEEEtran}
\usepackage{cite}

\ifCLASSINFOpdf
  \usepackage[pdftex]{graphicx}
  \graphicspath{{Figures/}}
  \DeclareGraphicsExtensions{.pdf,.jpeg,.png}
\else
  \usepackage[dvips]{graphicx}
  \graphicspath{{Figures}}
  \DeclareGraphicsExtensions{.eps}
\fi
\usepackage{amsmath}

\usepackage{algorithmic}
\usepackage{url}

\usepackage{diagbox}
\usepackage{hyperref}
\usepackage{pdflscape}
\usepackage{bm}
\usepackage{multirow}
\usepackage{color}
\usepackage{soul}
\sethlcolor{yellow}

\usepackage{listings}
\begin{document}
%

\title{Explanation Method for Anomaly Detection \\
on Mixed Numerical and Categorical Spaces}%

%
%

\author{Iñigo~López-Riobóo~Botana,
        Carlos~Eiras-Franco,
        Julio~Hernandez-Castro
        and~Amparo~Alonso-Betanzos
\thanks{Corresponding author: Iñigo López-Riobóo Botana \protect \\
E-mail: inigo.lopezrioboo.botana@udc.es}
\thanks{Iñigo López-Riobóo Botana, Carlos Eiras-Franco (e-mail: carlos.eiras.franco@udc.es) and Amparo Alonso-Betanzos (e-mail: ciamparo@udc.es) are with the Research Center on Information and Communication Technologies (CITIC) - Universidade da Coruña. Campus de Elviña, 15071 A Coruña, España.}
\thanks{Julio Hernandez-Castro is with the School of Computing - University of Kent. Cornwallis, CT2 7NF Canterbury, UK. e-mail: jch27@kent.ac.uk.}}

\markboth{Preprint version Computing Research Repository}
{López-Riobóo-Botana \MakeLowercase{\textit{et al.}}: Explanation Method for Anomaly Detection on Mixed Numerical and Categorical Spaces}

\maketitle

\begin{abstract}
Most proposals in the anomaly detection field focus exclusively on the detection stage, specially in the recent deep learning approaches. While providing highly accurate predictions, these models often lack  transparency, acting as ``black boxes". This criticism has grown to the point that explanation is now considered very relevant in terms of acceptability and reliability.
In this paper, we addressed this issue by inspecting the ADMNC (Anomaly Detection on Mixed Numerical and Categorical Spaces) model, an existing very accurate although opaque anomaly detector capable to operate with both numerical and categorical inputs. This work presents the extension EADMNC (Explainable Anomaly Detection on Mixed Numerical and Categorical spaces), which adds explainability to the predictions obtained with the original model. We preserved the scalability of the original method thanks to the Apache Spark framework.
EADMNC leverages the formulation of the previous ADMNC model to offer pre hoc and post hoc explainability, while maintaining the accuracy of the original architecture.
We present a pre hoc model that globally explains the outputs by segmenting input data into homogeneous groups, described with only a few variables. We designed a graphical representation based on regression trees, which supervisors can inspect to understand the differences between normal and anomalous data. Our post hoc explanations consist of a text-based template method that locally provides textual arguments supporting each detection.
We report experimental results on extensive real-world data, particularly in the domain of network intrusion detection. The usefulness of the explanations is assessed by theory analysis using expert knowledge in the network intrusion domain.
\end{abstract}


%
\IEEEpeerreviewmaketitle

\section{Introduction}
%
%
%
%
\IEEEPARstart{A}{n} anomaly is a pattern that is significantly different from the rest of the data. So much so that it arises suspicion that a different mechanism generated it \cite{hawkingsBook1}. Anomaly Detection is an old discipline in Statistics, also known as outlier detection \cite{Hodge04}. 
Over the last years, the field has grown considerably, as it has become particularly relevant in situations involving massive datasets where unexpected events may carry the most relevant information or have a great impact. 
These detection methods have found numerous applications in fields such as network intrusion detection \cite{kumarBook1}, surveillance \cite{vigilanciaPaper}, machinery monitoring \cite{Francos13} and many others.

\begin{figure}[!htb]
\centering
\includegraphics[height=2in]{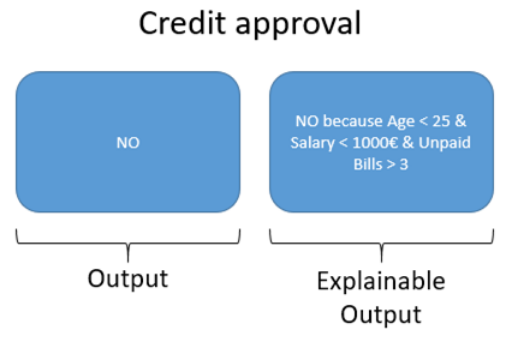}
\caption{Possible explanation for the outcome of an AI algorithm.} \label{eu_gdpr}
\end{figure}

Despite being capable, in some cases, of offering very effective detection rates, most anomaly detection algorithms cannot provide justifications for their outputs or extract new and actionable information carried in the data. This lack of explanation is, nowadays, one of the most critical shortcomings of Machine Learning \cite{explicacion_paper_XAI, Samek19}. DARPA (Defense Advanced Research Projects Agency) has a research line on XAI (Explainable Artificial Intelligence) \cite{darpa_xai, state-of-art_XAI}. The European Union cites explanation as one of the pillars of its Ethics Guidelines for Trustworthy Artificial Intelligence \cite{HLEGAI19}. These ideas make XAI an increasingly relevant notion, ``\textit{there is a need to find ways to explain the system to the decision-maker so that they know that their decisions are going to be reasonable}" \cite{state-of-art_XAI}.
These newly demanded capabilities are considered one of the most critical themes in the ``Third Wave of AI". Besides, regulatory measures such as EU's GDPR (General Data Protection Regulation) \cite{reglamento_general_proteccion_datos} are enforcing the use of explainable algorithms for decision making, particularly regarding citizens, as stated in the article 71 \cite{cita_reglamento_eu}.
This extra right to an explanation leads towards Artificial Intelligence algorithms that are more ethical and transparent, with each decision accompanied by a valid justification, avoiding ``black box" behaviour. Figure \ref{eu_gdpr} \cite{referencia_figuras_xai_eml} exemplifies the goal of these regulations.

There are two methods to attain explainability:
(1) Pre hoc explainability allows the supervisor to inspect the whole model and input data, while (2) post hoc explainability focuses on offering justifications for each single prediction made. The post hoc approach is, by far, the most prevalent strategy in explainable algorithms to date \cite{explicacion_paper_XAI, Samek-book}.

In general, in the machine learning community, there is no clear agreement between interpretation and explanation, and many times these terms are used interchangeably \cite{explanation_interpretation_definitions}. However, in \cite{rudin_explainability} a clear differentiation between inherently interpretable models and the explanation of black-box models is given. For this paper, we use the definition of Explainable Machine Learning that works on the idea of ``a second model created to explain the first black-box model"\cite{rudin_explainability}. This statement led us to explain the ADMNC black-box by training another surrogate model. In addition, explanation methods, both textual and visual, can be global (i.e., characterising the whole model and dataset) or local (i.e., explaining individual classification or regression outcomes)\cite{explanation_interpretation_definitions}. This paper explores both the global and local explanation methods on a mixed text-based and visual representation.

When discussing the explanation of black-box models like ours, it is crucial to consider the ``correlation does not imply causation" statement. In this way, we cannot deduce direct cause-and-effect relationships between the insights obtained with the new explanation module and the original detections made by the ADMNC algorithm \cite{admnc}. This criticism is already described in \cite{rudin_explainability}, claiming that explanations are often not reliable and can be misleading. However, these correlations are helpful to inspect the model, especially considering the contrast between normal and anomalous data. This turns out to be determining in generalisation power, detection of biases in the data or formulation of hypothesis for the abnormal data context, among others.

This work presents EADMNC (Explainable Anomaly Detection on Mixed Numerical and Categorical spaces), an explainable extension of the ADMNC (Anomaly Detector for Mixed Numerical and Categorical inputs) algorithm \cite{admnc}. The proposed method includes an extra layer that provides additional functionality capable of opening the ADMNC ``black box". Our contributions are as follows:

\begin{enumerate}
    \item We provide global explanations based on a pre hoc model by segmenting input data into homogeneous groups described with only a few variables. We designed a graphical representation based on regression trees, which supervisors can inspect to understand the differences between normal and anomalous data.
    \item We provide local post hoc explanations centred on a text-based template method, providing textual arguments supporting each detection. We leveraged the formulation of original ADMNC to find correlations between input features and detections.
    \item We offer supervisors a way for managing trade-offs between prediction precision and explanation complexity (through a tuned hyper-parameter and quality measure), handling both categorical and continuous features.
    \item We propose a set of metrics to measure the ``complexity" of the explanations.
    \item We propose a scalable algorithm that handles extensive real-world data, particularly in the domain of network intrusion detection. The algorithm is implemented using the Apache Spark framework to maintain the scalability of the original ADMNC system.
\end{enumerate}

The rest of this paper is structured as follows. Section \ref{sec:related} summarises the state-of-the-art in Explainable Anomaly Detection. Section \ref{sec:background} briefly describes the previous ADMNC algorithm \cite{admnc} and the ``black-box" problem. Section \ref{sec:proposed_method} presents the ``open-the-black-box" EADMNC scheme. In Section \ref{sec:experimentation} we detail the experiments performed to assess the validity of our approach. Lastly, Section \ref{sec:conclusions} lists the conclusions of this work and identifies future work paths.

\section{Related Work}\label{sec:related}


The ``open-the-black-box" approach proposes a surrogate model to explain an existing opaque or ``black-box" model. There are a set of well-known general frameworks \cite{review_general_explainable_surrogate} to provide these explanations and interpretations. LIME \cite{lime} is an explanatory framework that uses sparse linear models as explanations of any classifier \cite{lime_use_case_1_one_class_ae, lime_shap_both_use_case}. However, this framework does not handle parallelism and only considers locally faithful explanations of the predictions made. SHAP (SHapley Additive exPlanation) \cite{Shap} proposes both local and global explanations, assigning a score (i.e., SHAP value) to each input feature by looking at how the prediction changes when considering different feature subsets. However, the exact computation of SHAP values is challenging. This framework adds some assumptions to compute an approximate version of the SHAP values. It considers both missingness and feature independence. If the simplified inputs represent feature presence, then missingness requires features missing in the original input to have no impact \cite{Shap}. Moreover, model linearity is another optional assumption when computing the approximate SHAP values. SHAP framework has been extended to the tree-based models \cite{tree_shap}, proposing an accelerated GP-GPU version suitable for massively parallel computation \cite{gpu_tree_shap}. Many anomaly detection proposals \cite{shap_use_1, lime_shap_both_use_case, shap_diffi_use_case, shap_use_case_2, shap_use_case_3, shap_use_case_4} leverage SHAP to present explanations about their predictions. LRP (Layer-wise Relevance Propagation) \cite{lrp_original} is an explanation technique that can explain the predictions of complex state-of-the-art neural networks in terms of input features. This method propagates the prediction backwards in the neural network by means of propagation rules based on generalised linear mappings \cite{lrp_overview}. LRP also considers the non-linearity extension with local renormalisation layers \cite{lrp_extension}. Many deep neural networks for anomaly detection use LRP \cite{lrp_use_case_1, lrp_use_case_2} to provide explanations. However, this method is centred exclusively on the deep neural networks explanations. EXAD (EXplainable Anomaly Detection system) \cite{exad_framework} is an all-in-one complex anomaly detector that provides explanations using high-dimensional time series data.

In the anomaly detection field, we differentiate between the classic ``shallow" machine learning models and the emergence of new deep learning approaches \cite{deep_vs_classic_shallow_AD}. Explanations vary depending on the models, but also on the input data. Different types of data, such as image, text, time series data or numerical/categorical vectors entail many alternatives for anomaly explainability.

Focusing on the classic ``shallow" ML approaches, Quirk \cite{quirkExample} provides post hoc explanation over samples classified as anomalies, using conventional detection algorithms such as k-means or K-NN (k-Nearest Neighbors). Quirk's proposal lets the supervisor interact with the representation to obtain increasingly complex justifications. However, it fails to offer explanations and representations for data with more than three continuous dimensions and cannot process categorical features. LODI (Local Outlier Detection with Interpretation) \cite{LODI_explain}, LOGP (Local Outliers with Graph Projection) \cite{LOGP_explain} and \cite{explain_anomaly_general_methods} focus on feature selection and subspace separability, seeking for an optimal projection over input data to separate outliers from the neighbour inliers, constructing only local explanations with those subsets of variables. LODI does not handle large and high dimensional datasets, since its computation complexity is quadratic in the dimensionality. Moreover, it assumes an outlier can be linearly separated from inliers. LODI and LOGP do not handle categorical input data. Another recent work \cite{one_class_svm_rule_extractor} presents rule extractors over OCSVM (One-Class Support Vector Machines) to provide explanations for unsupervised anomaly detection. Although authors state their method is model-agnostic, the evaluation results are dependant on the two-kernel OCSVM model. Moreover, rule extractors are dependant on some assumptions they followed. DIFFI (Depth-based Isolation Forest Feature Importance) \cite{diffi_iforest_interpretation} and \cite{iforest_explanator_2} provide feature-based interpretation to the iForest anomaly detection algorithm \cite{iforest_anomaly_detection}. Both explanation methods were specifically designed for the iForest algorithm (i.e., they are not model-agnostic). Moreover, DIFFI method only provides global explanations, while in \cite{iforest_explanator_2}, only local explanations are presented. Finally, in \cite{lstm_gradient_boosted_decision_trees} gradient boosted ensemble of decision trees are used to provide local (but not global) explanations by feature importance.

Focusing on the novel deep learning approaches to provide explanations \cite{deep_alternatives_anomaly_detection, deep_explainable_anomaly_detection}, FCDD (Fully Convolutional Data Description) \cite{fcdd_heatmap_anomaly_explain} and \cite{deep_anomaly_heatmap_extraction} are examples of deep one-class classification models that include explanation heatmaps identifying the anomalous regions of the input images, but they are exclusively centred on image-based anomaly detection. In \cite{deep_cae_explain}, both CNN (Convolutional Neural Networks) and AE (AutoEncoders) models handle anomaly explanation over time series data, using the reconstruction errors to locally explain anomalies using association over previous similar anomalies. This model considers a set of explainable indicators defined by supervisors. GEE \cite{gee_vae_network_intrusion} is a framework designed for detecting and explaining anomalous traffic in NetFLow records. It leverages an unsupervised deep learning VAE (Variational AutoEncoder) model for detection and gradient-based fingerprints for local explanation. DevNet \cite{devnet_anomaly_scores_explanation} is a deep few-shot learning method that locally explains the anomalies by attributing the anomaly scores to the inputs through gradient back-propagation. It focuses exclusively on image datasets. The DAART (Detection of Anomalous Activity in Real-Time) system \cite{DAART_explain} relies on GAN (Generative Adversarial Network) and human-machine interactions to classify and explain detections. It provides the supervisor with post hoc examples and saliency maps to aid in image classification.

Our proposed EADMNC framework follows the CART (Classification And Regression Trees) \cite{book_CART_explained} approach. The global explanations are focused on regression trees that take into account both the anomaly scores of the opaque ADMNC model and the feature importance, following a pruning strategy led by a quality measure. Our graphical representation is simple and intuitive, making easier for the supervisor to inspect the whole model and input data under anomalous conditions. Moreover, the local explanations are centred on text-based templates to provide post hoc explanations over single detections, considering both the continuous and categorical features.

\section{Background}\label{sec:background}
The original ADMNC algorithm \cite{admnc} is a method developed for large-scale offline learning. It obtains a model of normal data that is then used to detect anomalies on mixed numerical and categorical spaces, working with two adjustable parametric models. We will consider a set of input data $D$ with both categorical and continuous variables:
\begin{equation}
D = \{(x_0,y_0),...,(x_{|D|},y_{|D|})\}.
\end{equation}
where $x_i$ is the continuous part and $y_i$ the categorical component of each instance. In order to treat each type of variable (numerical or categorical) differently, ADMNC uses the following heuristic factorisation of the pdf (probability density function):
\begin{equation}
\label{formula:general_pdf_factorization}
P(y,x) = P(y|x) P(x).
\end{equation}
With this approach, ADMNC can independently learn a model for each data type while accounting for any dependencies.

\subsection{Model for the numerical data}\label{sec:gmm}
ADMNC models continuous data employing a GMM (Gaussian Mixture Model), which is fit using Expectation-Maximization to maximise the likelihood of the training data \cite{EMgaussians}. The samples that the model deems unlikely can be, then, tagged as anomalies. GMM is well-known to be very dependant on initialisation values. To address that, a k-means clustering process is performed for a subset of the data, and the observed means and variances of the computed clusters are subsequently used.

\subsection{Model for the categorical data}\label{sec:logistic}
\label{section:categorical_formulas}
Given a categorical vector:
\begin{equation}
\label{formula:one_hot_representation}
\boldsymbol{y} = (y^0,...~,y^k), y^j \in \{0,1\},
\end{equation}
which can be obtained by One-Hot encoding \cite{oneHotCoding} the categorical variables, $P(y|x)$ is estimated as:
\begin{equation}
\label{form:general_lr}
P(\boldsymbol{y}|\boldsymbol{x}, \boldsymbol{w}) = \prod_{j = 0}^{k} P(Y = y^j|(\boldsymbol{x}, \boldsymbol{m}_j), \boldsymbol{w}),
\end{equation}
where $\boldsymbol{m}_j$ is the One-Hot encoding representation of $j$ and $\boldsymbol{w}$ the vector of parameters to be learned. Specifically, the probability of each vector component is estimated using:
\begin{equation}
\label{form:specific_lr_term}
P(Y = y^j|(\boldsymbol{x}, \boldsymbol{m}_j), \boldsymbol{w}) = \dfrac{1}{1 + e^{-(2y^j-1)\langle\boldsymbol{w},(\boldsymbol{x,m_j})\rangle}}.
\end{equation}
Where $\langle\boldsymbol{w},(\boldsymbol{x,m_j})\rangle$ is the dot product of the parameter vector $\boldsymbol{w}$ and $(\boldsymbol{x,m_j})$, which is the concatenated vector of continuous data $\boldsymbol{x}$, and masked representation $\boldsymbol{m}_j$. The resulting formula is analogous to that of Logistic Regression.

\subsection{Maximum likelihood parameter estimation}\label{sec:estimation}
Given a dataset $D$, ADMNC will find a set of parameters that maximizes the log-likelihood of the elements in $D$.
\begin{equation}
    log L(D) = \sum_{i=1}^{|D|}log P(\boldsymbol{y}_i|\boldsymbol{x}_i, \boldsymbol{w}) + \sum_{i=1}^{|D|} log P(\boldsymbol{x}_i).
\end{equation}
The factorisation of the probability estimation leads to independent terms for the log-likelihood that can be maximised separately or even simultaneously in parallel. In the case of the GMM used to model the numerical part of the data, the default implementation of GMM learning with Expectation-Maximization \cite{EMgaussians} included in the Apache Spark MLLib is used. As for the model used for $P(y|x)$, the probability estimation can be learnt using SGD (Stochastic Gradient Descent) \cite{estocasticGradientDescent}. The $\boldsymbol{w}$ vector is learnt using a mini-batch step \cite{mini_batch_step_gradient_desc} that can be parallelised thanks to the Apache Spark implementation. 
Since both models can be treated independently and each one offers parallelism, the scalability of the method (in terms of its effective use of additional computational resources to reduce processing time) is significantly increased. The use of several computing cores makes the training process much faster.

\subsection{The black box problem}\label{sec:anomaly_detection}
With ADMNC already trained in the normal data context, following the process described in Section \ref{sec:background}, we monitor new samples during the test, assigning them a continuous score based on the factorisation of the pdf from Formula \ref{formula:general_pdf_factorization}. Those samples whose score is below a pre-specified threshold (set by the model) will be considered anomalies \cite{admnc}.

However, we have neither justification nor explanation of why a specific continuous estimator is given to each new data sample (or why the estimator is lower or higher than the anomalous threshold). This process is entirely opaque, relies on the learned weights and is, thus, incomprehensible to end users.

\section{Proposed Method}\label{sec:proposed_method}
We propose the use of two different explanation methods to solve the ``black-box" problem of ADMNC. First, a regression tree global explanation will inform the supervisor of common features in anomalous samples that differentiate them from normal data.
Secondly, a post hoc local explanation of the factors that led to the detection of an anomaly will be generated, offering a text-based$^1$, intelligible description of the predictions made by ADMNC.

\subsection{Regression tree pre hoc explanation}\label{sec:prehoc_tree}
A single regression tree will use only the input variables to approximate the anomaly scores given by ADMNC. The leaf nodes of said tree define clusters that can be explained in terms of few input variables. This is a common technique when attempting to obtain explainable models \cite{darpa_xai_decision_trees, decision_trees_explanation_study}.
In order to build this tree, input samples will be ranked by ADMNC \cite{admnc} according to their anomaly score. These rankings will then be predicted from the original input variables using a regression CART model. Inspecting the coloured nodes of this model, following the gradient-based approach of the estimator from Figure \ref{fig:thresholds_explained}, the main features that lead data to be anomalous can be obtained.
In this model, the impurity measure (variance) indicates how disparate are the elements represented by a tree node. The goal of each split is to reduce this value as much as possible to better separate elements. Successive divisions make each node more representative of a specific ranking interval (i.e., of a certain anomaly level). By limiting the depth of the tree, we avoid overly complicated explanations, striking a balance between cluster homogeneity and explanation quality.

\subsubsection{Quality measure}\label{subsec:quality_measure}
To assess the fitness of the obtained explanations we have used a variation on the approach proposed in \cite{dyadic_data_explain}. We define the clustering $Cl(D)$ over dataset $D$ as a set of $m$ disjoint groups (equivalent to the number of clusters or leaf nodes) that contains every element in $D$. We can denote it as:
\begin{equation}
Cl(D) = \{Cl_1,...,Cl_m\}.
\end{equation}
Given a cluster $Cl_i$, its variance regarding the set of samples inside $N$ is given by:
\begin{equation}
\sigma^{2}_{~~Cl_i} = \dfrac{\sum_{i=1}^{|N|} (n_i - \mu)^2}{|N|},
\end{equation}
where $n_i$ corresponds to the ADMNC rank-based estimator for element $x_i$ in $Cl_i$ and $\mu$ is the mean of estimators for that cluster. To extend the formula to the whole clustering, the weighted variance (WV) of a $Cl(D)$ is defined as:
\begin{equation}
    WV(Cl(D)) = \dfrac{\sum\limits_{i \in 1..m} \sigma^{2}_{~~Cl_i}|Cl_i|}{|D|},
\end{equation}
where $m$ is the already known number of clusters in $Cl(D)$. The weighted variance of a clustering measures how homogeneous its components are and, consequently, its accuracy and predictive power.
In order to ensure that the clustering is easily understandable and explainable, this measure is complemented with another one that indicates the number of input variables employed to characterise each cluster $Cl_i$. As a result, the \emph{quality ($\mathcal{Q}$)} of a clustering is defined as:
\begin{equation}
    \label{formula:qualityClustering}
    \mathcal{Q}(CL(D)) = - WV(Cl(D)) - \lambda \sum_{Cl_i \in Cl(D)} NV(Cl_i),
\end{equation}
where $NV(Cl_i)$ represents the number of variables needed to describe the cluster $Cl_i$ and $\lambda$ is a hyper-parameter that allows the supervisor to balance the accuracy of the whole clustering with respect to the complexity of the explanation \cite{dyadic_data_explain}. This quality measure is always negative, and the goal is maximising its value by approaching 0. Maximising this measure will ensure that the groups obtained are as homogeneous as possible and explained using as few of the input variables as possible.

\subsubsection{Tree construction}
\label{subsubsec:tree_construction}
\begin{figure}[!htb]
\centering
\includegraphics[height=1.5in]{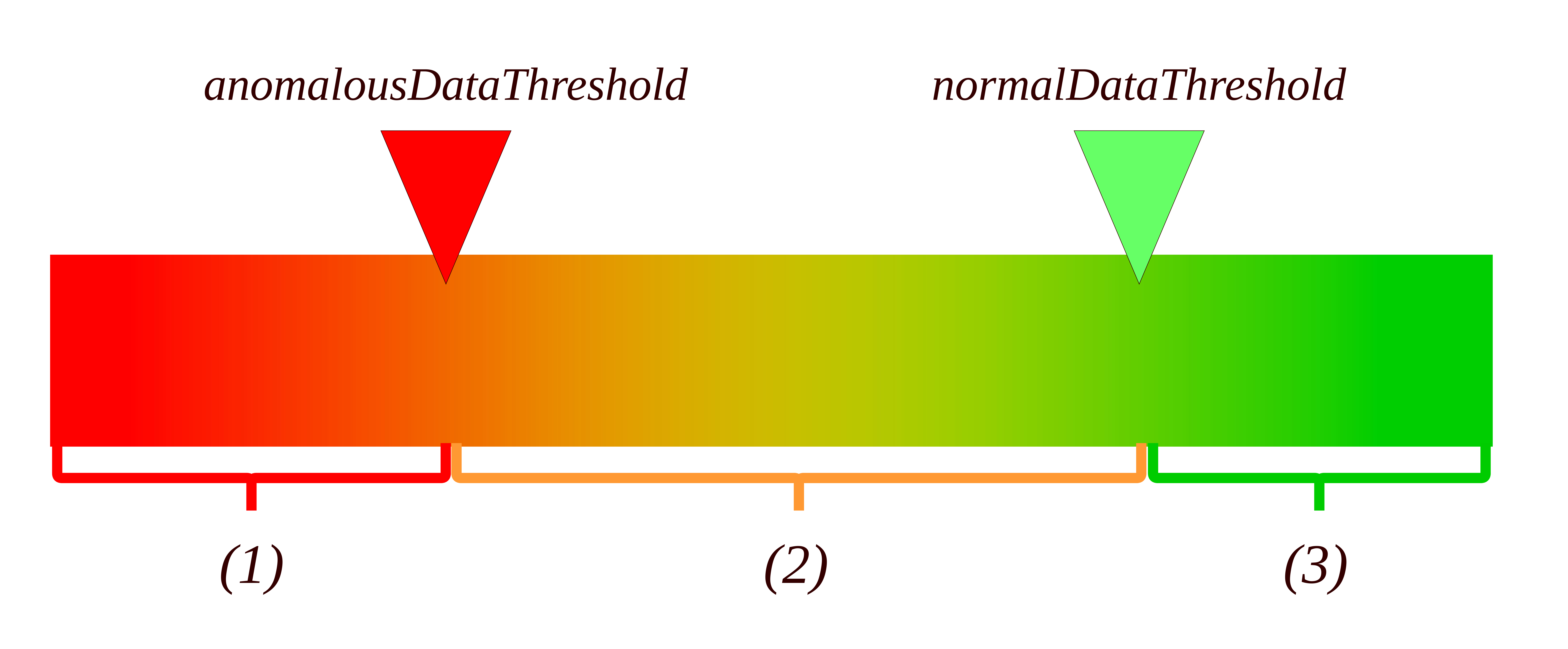}
\caption{The anomalous data range (1), the transition data range (2) and the normal data range (3). In order to focus on the anomalies, samples exceeding \textit{normalDataThreshold} (NDT) are assigned the same value so that the regression tree groups them in a single node of normal data. In this way, the latter sub-range (3) is shortened since we do not want to make an excessive effort in the division nor the explanation of normal data.} \label{fig:thresholds_explained}
\end{figure}

Having defined the measure used to assess clustering quality, the goal of our algorithm resides in finding the tree that maximizes the \textit{quality} $\mathcal{Q}$ of the whole clustering. We followed two different steps:

(1) first, a full $L_{MAX}$ level tree (with $L_{MAX}$ being a hyper-parameter) is built trying to minimise variance. This is achieved using the well-known CART \cite{book_CART_explained} algorithm employing variance as the impurity measure. The model is fit using the default Apache Spark MLLib implementation. Two thresholds (\textit{anomalousDataThreshold} and \textit{normalDataThreshold}) from previous ADMNC models are used to divide the estimator range $[0, 1]$ into up to three sub-ranges. Figure \ref{fig:thresholds_explained} shows the intuitive idea behind this.

(2) Subsequently, this full tree is pruned to optimise the quality measure. By doing so, node splits that decrease variance and quality are eliminated, yielding a more straightforward tree that maximizes quality. This pruning process is described in Figure \ref{alg:poda}. As a result, each leaf node corresponds to a separate group defined by the variables that lead to that node.

Implementing this algorithm in the Apache Spark framework allows us to perform most computations in parallel, since they are independent. The scalability of the method is guaranteed, allowing us to process very large datasets in manageable times, as we will show in Section \ref{sec:experimentation}.

\begin{figure}[!htb]
    \centering
    \begin{algorithmic}[1]
        \REQUIRE $tree$, $\lambda$.
        \ENSURE $Pruned~tree$.
        \STATE \textbf{function} \textsc{prune}($tree$,$\lambda$) $\rightarrow$ $tree$
        \begin{ALC@g}
        \IF{\textsc{isLeaf}($tree$)}
        \RETURN $tree$;
        \ENDIF
        \STATE $left \leftarrow $\textsc{prune}$(tree.leftBranch,\lambda)$;
        \STATE $right \leftarrow $\textsc{prune}$(tree.rightBranch,\lambda)$;
        \STATE $splitVariance\leftarrow$ \\ $\frac{left.variance*|left|+right.variance*|right|}{|tree|}$;
        \STATE $\Delta E=splitVariance-tree.variance$;
        \STATE $\Delta NV=$\\$left.numVars+right.numVars-tree.numVars$;
        \IF{$-\Delta E-\lambda\Delta NV <= 0$}
        \STATE $tree.leftBranch \leftarrow \emptyset$;
        \STATE $tree.rightBranch \leftarrow \emptyset$;
        \ENDIF
        \RETURN $tree$;
        \end{ALC@g}
        \end{algorithmic}
    \caption{Pruning algorithm.}
    \label{alg:poda}
\end{figure}

\subsection{Tree transcription text-based post hoc explanation}\label{sec:post_hoc_tree}
In most cases, an anomalous sample will be classified by the regression tree in a node describing anomalous data, corresponding to the first sub-range from Figure \ref{fig:thresholds_explained}. Therefore, the first post hoc explanation is centred on the transcription of the path to that node in the tree.

In this way, given the subset of the top $N$ (set with a hyper-parameter) anomalies detected by ADMNC, these explanations are presented as follows$^2$:

\begin{lstlisting}[language=bash, keywords={smooth}, basicstyle=\small\ttfamily, mathescape=true]
Positive anomaly detection N(1).
* Explanation: 
$\longrightarrow$ Feature "Down/Up Ratio" <= 0.5
$\longrightarrow$ Feature "Flow IAT Mean" <= 13729.499
$\longrightarrow$ Feature "Flow Duration" <= 940.0
$\longrightarrow$ Feature "TotLen Bwd Pkts" <= 123.5
$\longrightarrow$ Feature "Flow IAT Max" <= 602.5 
* These features place the item in a cluster containing 1.104% of all elements.
\end{lstlisting}

This transcription determines the correlation between input variables and the anomalous condition, providing information about the proportion of data inside the anomalous cluster.

\subsection{The other text-based post hoc explanations}\label{sec:posthoc_rules}
For the rest of the cases (i.e., when the regression tree does not lead anomalous samples to nodes describing anomalous data), the formulation of ADMNC described in Section \ref{sec:background} is examined to determine the relevant factors that led to the detection. For this approach, (1) the continuous part is inspected using both the GMM model \cite{admnc} and the regression tree, integrating them on a simple Rule-based classification System. For the categorical part, (2) the explanation is focused on the analysis of the Logistic Regression model Formulas \ref{form:general_lr} and \ref{form:specific_lr_term}.

\subsubsection{Rule-based classification system}
The text-based post hoc explanations of this alternative will consist of arguments identifying the following correlations between the anomalous detection and the continuous input data:
\begin{enumerate}
    \item Continuous sample $\boldsymbol{x_i}$ has a pdf value lower than a pre-specified threshold (this being a hyper-parameter) for all Gaussians in the GMM. That means the probability of belonging to any of them is very low. Therefore, the sample does not fit normal data.
    \item Continuous sample $\boldsymbol{x_i}$ is assigned to a Gaussian that represents a tiny fraction of samples. Therefore $\boldsymbol{x_i}$ is sufficiently infrequent and can be considered as an anomaly.
    
    

\end{enumerate}

\subsubsection{Logistic regression analysis}
The text-based post hoc explanations of this alternative will consist of arguments identifying the correlations between the anomalous detection and the input categorical data. In this case, the explanation will state that one or more categorical input values from $y$ are infrequent and, therefore, correlated with the anomaly detected. To identify these values, we process all categorical estimators from Formula \ref{form:general_lr} and keep these values as pairs of (categorical estimator, categorical term $y^j$) on a list called $E$, ordered from lower to higher estimator values as follows:
\begin{equation}
\begin{split}
E = & [(est_{y^0}, y^0),...,(est_{y^k}, y^k)] ~/~ \\
& (est_{y^j} <= est_{y^{j+1}}).
\end{split}
\end{equation}


To determine a suitable subset of $E$, we apply a simple threshold filtering ($T_{filter}$) to obtain a processed list $E'$. The goal is to identify the $y_j$ terms in Formula \ref{form:general_lr} that drive $P(y|x,w)$ more significantly down. In other words, we look for One-Hot encoding terms with low categorical estimators, since they identify the variables of the input vector that are the source of that particular detection.

We also aim at retrieving the relation of categorical terms with respect to continuous data, given the conditional probability function shown in Formula \ref{formula:general_pdf_factorization}. For each previously selected categorical term $y^j$ of list $E'$, we focus on the dot product $\langle\boldsymbol{w},(\boldsymbol{x,m_j})\rangle$ from Formula \ref{form:specific_lr_term} to look for the addend that makes the categorical estimator lower. If that addend corresponds to the continuous part $x_i$ of $(\boldsymbol{x_i,m_j})$, we can justify the relation between categorical and continuous data, based on conditional probability.

\subsubsection{Description of combined post hoc explanation}

This mixed approach for textual explanations using both the (1) Rule-based Classification System for the continuous part of the data and (2) the Logistic Regression analysis for the categorical part is presented in an HTML (Hypertext Markup Language) report. An example of anomaly detection described in this report is presented next$^3$, summarised in a brief human-readable text. All the information presented here is further detailed in Appendix \ref{appendix:detailed_post_hoc_explanation}.

\begin{lstlisting}[language=bash, keywords={smooth}, basicstyle=\small\ttfamily, mathescape=true]
Detected anomaly N(6):
* Explanation: 
$\rightarrow$ (1) It is an anomaly since the continuous sample is clearly separated from learned groups.
$\rightarrow$ (2) The model considers improbable that a normal sample could have in the categorical feature "Existence of ragged T wave" a value of 0.0
$\rightarrow$ (2) The model considers improbable that a normal sample could have in the categorical feature "Existence of ragged P wave" a value of 0.0 knowing the continuous value of -0.525 in the feature "channel_V5_amplitude_Rwave".
\end{lstlisting}

\begin{figure}[!htb]
\centering
\includegraphics[height=1.5in]{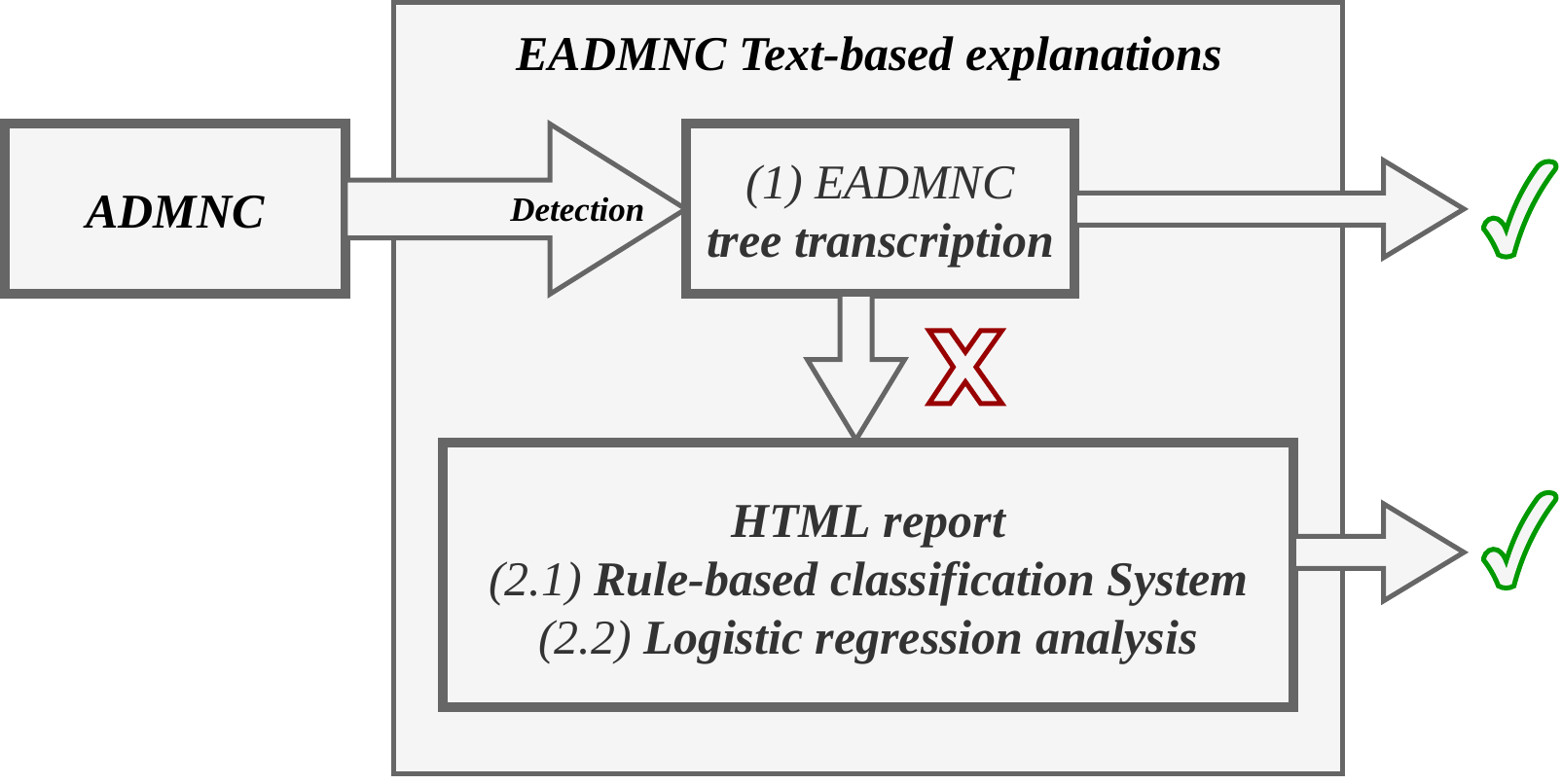}
\caption{Text-based explanation process given an anomalous detection. These results are provided through regression tree transcriptions (1) and by means of the mentioned HTML report (2.1 and 2.2).} \label{fig:text-based-explanation-process}
\end{figure}

\begin{figure}[!htb]
\centering
\includegraphics[width=3.85in]{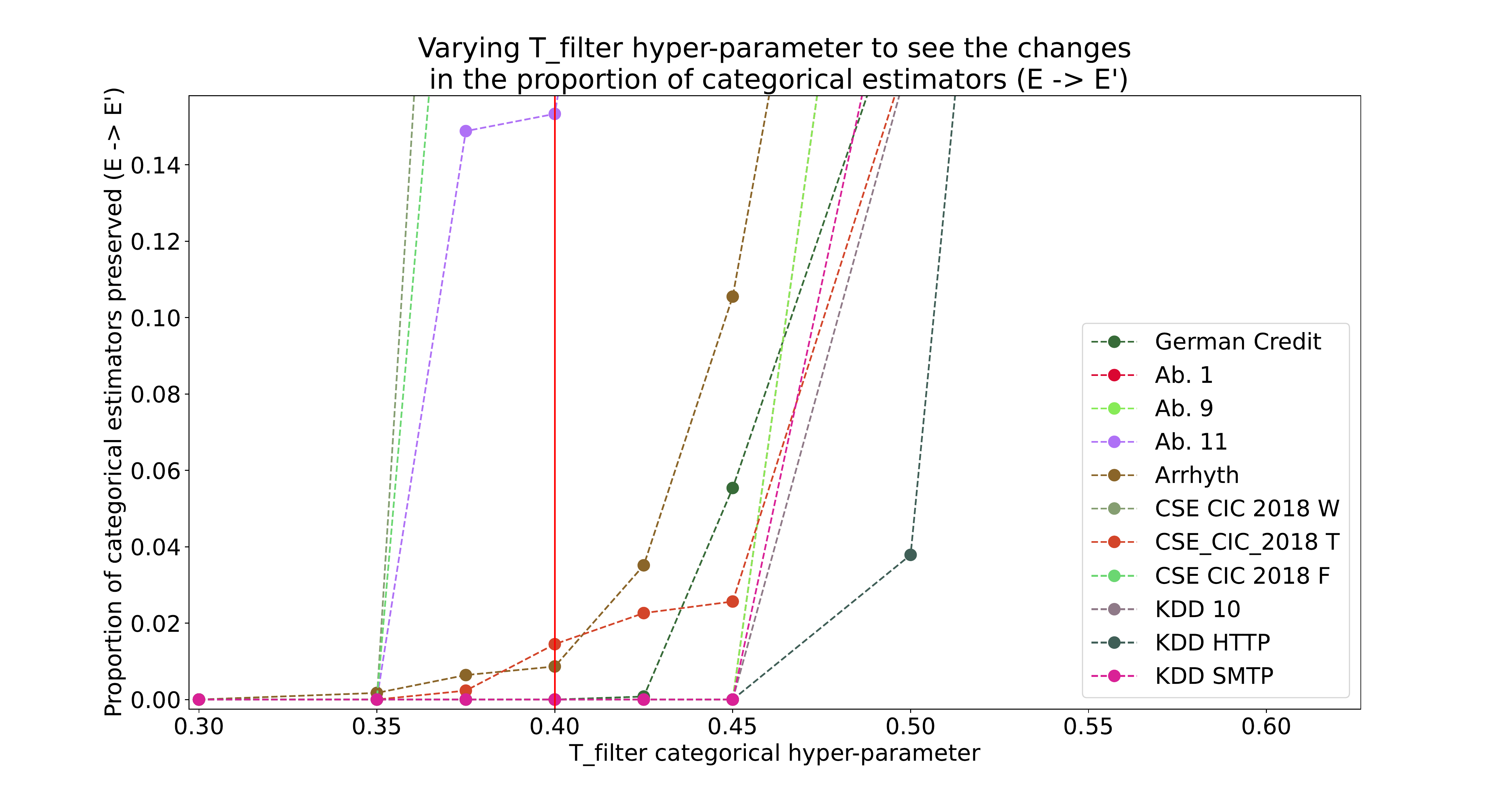}
\caption{Proportion of categorical terms preserved ($E \rightarrow E'$) changing $T_{filter}$ hyper-parameter. Tests were made over 5 executions of each dataset. The higher the $T_{filter}$, the more terms are preserved. We finally set the value to 0.4, a threshold from which the lowest categorical estimators started to be detected in several datasets.}
\label{fig:categorical_hyper_parameters}
\end{figure}

Figure \ref{fig:text-based-explanation-process} summarises the text-based explanation process given an anomalous detection and Table \ref{table:post_hoc_hyperparameter_description} shows values set for the required hyper-parameters. Experts can customise them to adjust the complexity of the text-based explanations (i.e., the level of detail). On the one hand, $L_{MAX}$ and \textit{number of bins} were set accordingly to simplify paths and node splits along the trees. On the other hand, the trade-off configuring the categorical hyper-parameter $T_{filter}$ depends on the detection sensitivity related to the lowest categorical estimators (and therefore, to the most infrequent categorical terms $y^j$). Figure \ref{fig:categorical_hyper_parameters} depicts the idea. In a few words, we selected the hyper-parameter value that enabled us to choose the most infrequent categorical terms$^4$ $y^j$. Then, with this subset of terms, we analysed the correlations with the anomalous condition.

\begin{table}[!htb]
\caption{Values of the hyper-parameters used for the text-based explanations. The first subset is related to the tree transcription, whereas the second subset is related to the logistic regression analysis.}
\label{table:post_hoc_hyperparameter_description}
\centering
\begin{tabular}{|ll|c|}
\hline
Regression tree $L_{MAX}$ & \extracolsep{\fill} & \multicolumn{1}{c|}{5} \\
Regression tree number of bins & \extracolsep{\fill} & \multicolumn{1}{c|}{40} \\



\cline{1-3}
Categorical explanation T$_{filter}$ & \extracolsep{\fill} & \multicolumn{1}{c|}{0.40} \\
\hline
\end{tabular}
\end{table}

\section{Experimental Results}\label{sec:experimentation}

In this section, we present the experiments performed to assess the validity of our approach. 
We have used the same methodology as in the original ADMNC paper \cite{admnc}, which consists in employing both real-world and synthetic datasets. Namely, we have used the same five small datasets from the previous work \cite{admnc}, extracted and adapted from the UCI Machine Learning Repository \cite{BBDD_datasets} (Arrhythmia, German Credit, and three versions of Abalone). We also considered the same two large datasets \cite{admnc} focusing on the network intrusion detection domain: KDDCup99 \cite{kdd-cup-99} and ISCXIDS 2012 \cite{ids-2012}. Additionally, we included a new dataset named CSE-CIC-IDS 2018 \cite{ids-2018} (an improved and more complete version of the previous ISCXIDS 2012). These three datasets were used to verify the validity of our method for large-scale data. Finally, we used a synthetic dataset generator \cite{admnc} to test our algorithm using different numbers of variables with abnormal values as a proxy for detection difficulty. The more abnormal the values, the easier the anomaly detection should be.

\begin{table}[!htb]
\caption{Complexity metrics for explanations. Metrics between brackets stand for the values before pruning the trees, whereas outer metrics correspond to the values after pruning. Complexity is summarised by means of $\#Cl$ (number of clusters) and NV (number of variables to reach all clusters), whereas WV (weighted variance) opposes the homogeneity of the clusters. Estimators are within the range $[0, \text{NDT}]$.}
\label{table:summary_results}
\centering
\begin{tabular}{|l|c|c|c|}
\hline
\textbf{Dataset} & \multicolumn{3}{|c|}{\textbf{Complexity of explanation}} \\ \hline
\textbf{Name} & \textbf{WV} & \boldsymbol{$\#Cl$} & \textbf{NV} \\ \hline
Arrhythmia & 0.012(0.010) & 6(24) & 17(113) \\ \hline
German Credit & 0.003(0.001) & 5(26) & 12(126) \\ \hline
Ab. 1 & 0.007(0.007) & 11(32) & 41(160) \\ \hline
Ab. 9 & 0.005(0.004) & 11(32) & 41(160) \\ \hline
Ab. 11 & 0.005(0.005) & 13(31) & 52(154) \\ \hline
IDS-2012 & $6.54\times10^{-6}$($6.45\times10^{-6}$) & 22(32) & 102(160) \\ \hline
IDS-2018-F & 0.003(0.002) & 10(32) & 36(160) \\ \hline
IDS-2018-T & $1.22\times10^{-5}$($9.63\times10^{-6}$) & 11(32) & 42(160) \\ \hline
IDS-2018-W & 0.002(0.002) & 14(32) & 58(160) \\ \hline
KDD-FULL & 0.011(0.011) & 13(30) & 53(148) \\ \hline
KDD-SMTP & $7.08\times10^{-7}$($5.23\times10^{-7}$) & 6(32) & 17(160) \\ \hline
KDD-HTTP & $1.27\times10^{-7}$($7.46\times10^{-8}$) & 3(32) & 5(160) \\ \hline
KDD-10 & $3.03\times10^{-6}$($2.36\times10^{-6}$) & 6(31) & 17(154) \\ \hline
Synth-1 NV & 0.0002(0.0001) & 3(26) & 5(126) \\ \hline
Synth-2 NV & 0.0001(0.0001) & 5(26) & 14(125) \\ \hline
Synth-4 NV & 0.0002(0.0001) & 5(30) & 14(148) \\ \hline
Synth-8 NV & 0.0004(0.0003) & 7(27) & 21(130) \\ \hline
Synth-16 NV & 0.0003(0.0002) & 11(28) & 41(137) \\ \hline
Synth-32 NV & 0.0003(0.0002) & 10(29) & 36(143) \\ \hline
\end{tabular}
\end{table}

Comparisons with other anomaly detection methods are omitted in this work since ADMNC has already been compared to a set of well-known state-of-the-art algorithms \cite{admnc}, and the extensions included for explainability in EADMNC do not affect detection accuracy.

To be able to work with large datasets$^5$ such as KDDCup99, ISCXIDS 2012 and CSE-CIC-IDS 2018, some steps were taken to pre-process and clean the data. We adapted the network intrusion detection domain to the anomaly detection task for these three datasets, assuming that attacks of any class are anomalies. For KDDCup99, we built up three additional variants by transforming the original dataset: KDDCup99 (10\%) is the reduced version which contains only 10\% of the instances; KDDCup99 (HTTP) is the result of filtering the dataset to keep only HTTP connections, and, analogously, KDDCup99 (SMTP) only contains SMTP activity. The detailed process for obtaining these datasets is explained in our previous work on the ADMNC algorithm \cite{admnc}.
On the other hand, we handled ISCXIDS 2012, performing a common pre-processing step and data cleaning. Lastly, CSE-CIC-IDS 2018 was used on different attack contexts, regarding several daily logs as provided by the research group \cite{ids-2018-web} that generated it. In this case, one dataset per weekday was chosen to carry out the following experimentation. We selected \textit{friday-16-02-2018}, \textit{thursday-15-02-2018} and \textit{wednesday-21-02-2018}. Each one of them contains more than a million samples. We adapted them, so all benign traffic was identified as non-anomalous. Every type of network attack (DoS attacks-Hulk, DoS attacks-SlowHTTPTest, DoS attacks-Slowloris, etc.) was flagged as anomalous.

The learning of the anomaly detection model was performed using the same method as in \cite{admnc}, working with 70\% of data for training and 30\% for testing, and performing five repetitions for each test. The effectiveness of the detection is measured with the average AUROC (Area Under the Receiver Operating Characteristic) \cite{auroc_Melo2013} of these executions. Standard deviations are listed to attest to the stability of the method. The supervisor receiving the explanations should first consider the accuracy of the anomaly detection model to get a context of how valid the explanations are. Since the explanation layer opens the ADMNC ``black box", it can not obtain relevant information regarding the data if the underlying anomaly detection model does not fit it adequately.

We examined the results of the pre hoc explanation in terms of complexity, making a comparison between full and pruned metrics for each dataset. Results are listed in Table \ref{table:summary_results} ($\mathcal{Q}$ and $\lambda$ values are available in Appendix \ref{appendix:table_q_lambda}). Pruning algorithm from Figure \ref{alg:poda}, applying pre-specified $\lambda$ value, maximizes negative $\mathcal{Q}$ (quality) value to approach zero. That means the best balance between accuracy (dependent on WV, the weighted variance, which is opposed to the homogeneity of the clusters) and complexity of explanation (dependent on NV, the number of variables to reach the clusters). The weighted variance (WV) defined in Section \ref{subsec:quality_measure} is propagated from leaf nodes to the upper levels until reaching the root node. We can see that reducing the complexity of the explanation (i.e., deleting superfluous clusters and therefore reducing NV) also increases WV (homogeneity loss) due to the disposal of related branches. This is an important trade-off between the accuracy and complexity of the explanation.

We set hyper-parameter $\lambda$ and increase or decrease its value accordingly with the pruning effort. Higher values for $\lambda$ mean extensive work on pruning branches since the quality Formula \ref{formula:qualityClustering} obtains lower values that must be close to zero after pruning. In comparison, lower values for $\lambda$ indicate moderate pruning. The $\lambda$ value can be modified by the supervisor as a configuration value, assigning more or less importance to the complexity of explanation with regards to accuracy. To set these values in our experiments, we sought pruned trees with the best anomalous context representations (i.e., with anomalous clusters that correspond the best to the input anomaly ratio of the dataset). Final $\lambda$ values are available in Appendix \ref{appendix:table_q_lambda}.

Examples of explanatory trees are shown in Figures \ref{fig:cse-cic-2018-thursday-paper} and \ref{fig:cse-cic-2018-friday-paper}. The algorithm provides paths to explain the anomalous and transition data and preserves almost all the normal data in the same node. All the pre hoc and post hoc explanations of the listed experiments are available in a public online repository$^6$. It is worth mentioning that, as stated in Section \ref{subsubsec:tree_construction}, the estimator range $[0, 1]$ is shortened using the \textit{normalDataThreshold} (NDT) parameter, as already depicted in Figure \ref{fig:thresholds_explained}. Thus, metrics provided in Table \ref{table:summary_results} correspond to the estimator range $[0, \text{NDT}]$. This parameter (NDT) can be examined in Table \ref{result:pre-hoc-mse-normaldataThreshold}. Looking at the examples in Figure \ref{fig:cse-cic-2018-thursday-paper} and \ref{fig:cse-cic-2018-friday-paper}, we can observe that some clusters were deleted (shadowed) by the pruning algorithm from Figure \ref{alg:poda}. In this way, the number of input variables required by the explanation ($NV$) is decreased while preserving as much of the homogeneity of the clusters as possible.

\begin{table}[!htb]
\caption{MSE (Mean Squared Error) as fitness function for regression trees. Estimators are in the range $[0, NDT]$ and $NDT \in (0, 1]$.}
\label{result:pre-hoc-mse-normaldataThreshold}
\centering
\begin{tabular}{|l|c|c|}
\hline
\rule{0\normalbaselineskip}{1.5\normalbaselineskip}
\textbf{Dataset} & \textbf{MSE} & \textbf{NDT} \\
& $\bm{(\mu \pm \sigma)}$ & \\
[1.5 mm] \hline
\rule{0\normalbaselineskip}{1\normalbaselineskip}
Arrhythmia & $0.0330$ $\pm$ $0.0050$ & 0.479 \\ [1 mm] \hline
\rule{0\normalbaselineskip}{1\normalbaselineskip}
German Credit & $ 0.0048$ $\pm$ $0.0019$ & 0.305 \\ [1 mm] \hline
\rule{0\normalbaselineskip}{1\normalbaselineskip}
Ab. 1 & $0.0071$ $\pm$ $0.0001$ & 0.346 \\
~Ab. 9 & $0.0050$ $\pm$ $0.0003$ & 0.318 \\
~Ab. 11 & $0.0057$ $\pm$ $0.0002$ & 0.346 \\ [1 mm] \hline
\rule{0\normalbaselineskip}{1\normalbaselineskip}
IDS-2012 & $1.03\times10^{-5}$ $\pm$ $2.72\times10^{-6}$ & 0.051 \\ [1 mm] \hline
\rule{0\normalbaselineskip}{1\normalbaselineskip}
IDS-2018-F & $0.0027$ $\pm$ $0.0008$ & 0.615 \\
~IDS-2018-T & $1.14\times10^{-5}$ $\pm$ $1.46\times10^{-6}$ & 0.046 \\
~IDS-2018-W & $0.0008$ $\pm$ $0.0004$ & 0.565 \\ [1 mm] \hline
\rule{0\normalbaselineskip}{1\normalbaselineskip}
KDD-FULL  & $0.0109$ $\pm$ $0.0005$ & 0.865 \\
~KDD-SMTP  & $6.44\times10^{-7}$ $\pm$ $1.17\times10^{-7}$ & 0.018 \\
~KDD-HTTP  & $1.30\times10^{-7}$ $\pm$ $5.1\times10^{-8}$ & 0.010 \\
~KDD-10  & $4.07\times10^{-6}$ $\pm$ $5.31\times10^{-7}$ & 0.033 \\ [1 mm] \hline
\rule{0\normalbaselineskip}{1\normalbaselineskip}
Synth-1 NV  & $0.0001$ $\pm$ $2.98\times10^{-5}$ &  0.081 \\
~Synth-2 NV  & $0.0001$ $\pm$ $3.09\times10^{-5}$ & 0.077 \\
~Synth-4 NV  & $0.0001$ $\pm$ $2.37\times10^{-5}$ & 0.084 \\
~Synth-8 NV  & $0.0004$ $\pm$ $5.10\times10^{-5}$ & 0.109 \\
~Synth-16 NV & $0.0004$ $\pm$ $2.47\times10^{-5}$ & 0.100 \\
~Synth-32 NV & $0.0004$ $\pm$ $3.71\times10^{-5}$ & 0.105 \\ [1 mm] \hline
\end{tabular}
\end{table}

We have measured the completeness of both text-based methods as the fraction of detections that received a textual explanation, as shown in Table \ref{result:fraction_post_hoc_explain}.

\begin{table}[!htb]
\caption{Fraction of anomalies with text-based post hoc explanations (adjustable parameter set to the first 400 anomalies). AUROC is the fitness function for anomaly detection. Both fractions (the tree transcriptions (1) and the HTML report (2)) are shown next to each other separated by a hyphen. The previous ADMNC model must fit properly to detect anomalies (AUROC).}
\label{result:fraction_post_hoc_explain}
\centering
\begin{tabular}{|l|c|c|}
\hline
\rule{0\normalbaselineskip}{1.5\normalbaselineskip}
\textbf{Dataset} & \textbf{AUROC} & \textbf{Fraction of explanations} \\
& $\bm{(\mu \pm \sigma)}$ & \\
[1.5 mm] \hline
\rule{0\normalbaselineskip}{1\normalbaselineskip}
Arrhythmia & 0.792 $\pm$ 0.050 & 1.000 - 1.000 \\ [1 mm] \hline
\rule{0\normalbaselineskip}{1\normalbaselineskip}
German Credit & 0.620 $\pm$ 0.040 & 0.000 - 0.986 \\ [1 mm] \hline
\rule{0\normalbaselineskip}{1\normalbaselineskip}
Ab. 1 & 0.794 $\pm$ 0.020 & 1.000 - 0.220 \\
~Ab. 9 & 0.588 $\pm$ 0.020 & 1.000 - 0.250 \\
~Ab. 11 & 0.792 $\pm$ 0.010 & 1.000 - 0.118 \\ [1 mm] \hline
\rule{0\normalbaselineskip}{1\normalbaselineskip}
IDS-2012 & 0.919 $\pm$ 0.020 & 0.000 - 1.000 \\ [1 mm] \hline
\rule{0\normalbaselineskip}{1\normalbaselineskip}
IDS-2018-F & 0.938 $\pm$ 0.013 & 1.000 - 1.000 \\
~IDS-2018-T & 0.829 $\pm$ 0.021 & 0.000 - 1.000 \\
~IDS-2018-W & 0.955 $\pm$ 0.014 & 1.000 - 1.000 \\ [1 mm] \hline
\rule{0\normalbaselineskip}{1\normalbaselineskip}
KDD-FULL  & 0.939 $\pm$ 0.011 & 1.000 - 1.000 \\
~KDD-SMTP  & 0.980 $\pm$ 0.010 & 0.643 - 1.000 \\
~KDD-HTTP  & 0.992 $\pm$ 0.010 & 1.000 - 1.000 \\
~KDD-10  & 0.966 $\pm$ 0.020 & 0.548 - 1.000 \\ [1 mm] \hline
\rule{0\normalbaselineskip}{1\normalbaselineskip}
Synth-1 NV  & 0.614 $\pm$ 0.050 & 1.000 - 0.135 \\
~Synth-2 NV  & 0.743 $\pm$ 0.020 & 1.000 - 0.161 \\
~Synth-4 NV  & 0.837 $\pm$ 0.020 & 1.000 - 0.146 \\
~Synth-8 NV  & 0.958 $\pm$ 0.010 & 1.000 - 0.107 \\
~Synth-16 NV & 0.998 $\pm$ 0.000 & 1.000 - 0.137 \\
~Synth-32 NV & 1.000 $\pm$ 0.000 & 0.502 - 0.289 \\ [1 mm] \hline
\end{tabular}
\end{table}

\subsection{Scalability tests}
We performed our experiments with the Apache Spark distributed framework, using a computer cluster formed by 8 machines (called executors) with 12 computing cores each. Technical specifications for each node are provided in Table \ref{table:Cesga_overview}. We configured different scenarios, varying the number of machines to observe the changes in performance and execution time. Scalability allows us to proportionally increase the speed of the algorithms as we configure more machines in the cluster. We can achieve this because the Apache Spark architecture provides a way to distribute data partitions over different executors, responsible of specific tasks. Each worker node is responsible for its computation, and partial results are sent to be gathered together. The driver node carries this process and retrieves the final results. Table \ref{result:scalability_test} summarises our tests using the original full KDDCup99 dataset, with almost five million samples. We can see that the ADMNC training phase is the most expensive in time units. In contrast, considering the training phase of the regression tree and the explanation process, the new explainable layer does not constitute a time-consuming task. Changes in execution time varying the number of executors can be slightly disproportionate with respect to the theoretical desired behaviour due to issues related to experimentation (e.g., shared limited memory resources cause garbage collector calls and fault tolerance of nodes in Spark involves task recovery and extra time consumption).

\begin{table}[!htb]
\caption{CESGA (Centro de Supercomputación de Galicia) computer cluster overview}
\label{table:Cesga_overview}
\centering
\begin{tabular}{|l c|}
\hline
\multicolumn{2}{|c|}{\textbf{8 nodes with the following characteristics:}}\\ \hline
\textbf{Processor}: & 2 x Intel Xeon E5-2620 v3 at 2.40Ghz  \\
\textbf{Cores:} & 6 per processor (12 per node)  \\ 
\textbf{Threads:} & 2 per core (24 total per node)  \\ 
\textbf{Storage:} & 12 x 2TB NL SATA 6Gbps 3.5” G2HS  \\ 
\textbf{RAM:} & 64 GB  \\ 
\textbf{Network:} & 1x10Gbps + 2x1Gbps  \\ \hline
\end{tabular}
\end{table}

\begin{table}[!htb]
\caption{Scalability tests (time unit in minutes) using 2, 4 and 8 executors with original KDDCup99 full dataset. *ADMNC only considers original ADMNC algorithm, while E* only considers the new explainable layer.}
\label{result:scalability_test}
\centering
\begin{tabular}{|l|l|c|c|c|c|}
\hline
\textbf{Stage} & \textbf{Task description} & \multicolumn{3}{|c|}{\textbf{\#~Executors}} \\ \hline
 & & \textbf{2} & \textbf{4} & \textbf{8} \\ 
 \textbf{Anomaly detection} & \textbf{*ADMNC training} & 5.344 & 2.554 & 1.688 \\ \hline
 \multirow{2}{*}{\textbf{Explainable layer}} & \textbf{E* training} & 1.125 & 0.687 & 0.450 \\
 & \textbf{E* explanation} & 0.636 & 0.368 & 0.229 \\ \hline
 \textbf{Full process} & \textbf{EADMNC algorithm} & 7.105 & 3.609 & 2.367\\ \hline
\end{tabular}
\end{table}

\subsection{Explanation of the results in the network intrusion detection domain}\label{sec:analysis_explanations}

Assessing the usefulness of the results that EADMNC provides is a complicated issue due to (1) the need for expertise in the domain and (2) the subjective nature of the concept of explanation usefulness.
Other than using a quantitative measure such as the Quality measure proposed in \cite{dyadic_data_explain}, only the analysis given by experienced users (i.e., their interpretations) can provide us valuable feedback on their usefulness.
To that effect, we report a study (further detailed in Appendix \ref{appendix:qualitative_assessment_trees}) of the trees found from the point of view of network intrusion detection. It shows that the obtained representations enable us to provide explanations that are both coherent with the existing knowledge in the field and insightful.

\begin{landscape}
\centering
\begin{figure}
\centering
\includegraphics[width=\linewidth, height=0.30\textwidth]{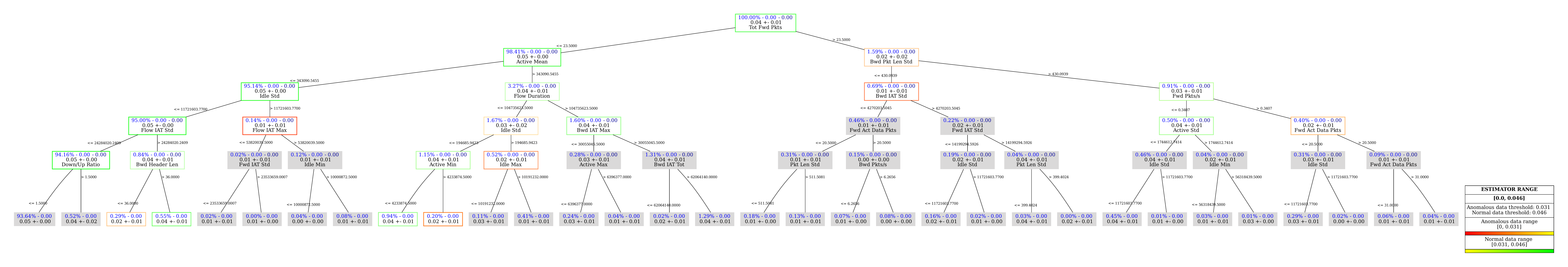}
\caption{Explanatory tree before and after pruning ($\lambda = 10^{-7}$) for the CSE-CIC-IDS 2018-thursday-15 dataset. Named sequentially, reading from left to right, each node shows: the proportion of elements that it represents w.r.t. the full dataset (in blue), overall variance at node (blue), the weighted variance at that node w.r.t children nodes (dark blue). Mean and standard deviation for the subset of estimators on each node are also provided. Split values are shown next to each split line with their features names, included at the bottom of each non-leaf node. Pruned clusters using Formula \ref{formula:qualityClustering} are shadowed. Summary table gives information in relation to selected thresholds and colour gradients.}
\label{fig:cse-cic-2018-thursday-paper}
\end{figure}
\begin{figure}
\centering
\includegraphics[width=\linewidth, height=0.30\textwidth]{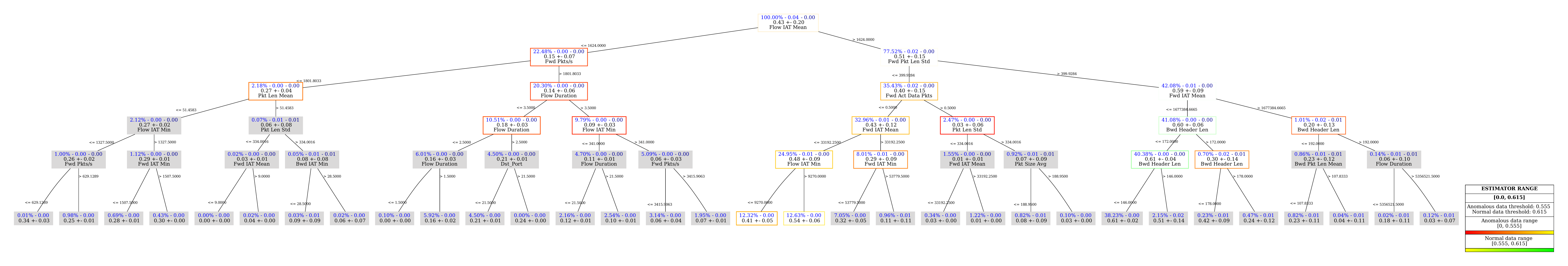}
\caption{Explanatory tree before and after pruning ($\lambda = 10^{-4}$) for the CSE-CIC-IDS 2018-friday-16 dataset. Information given in this tree is identical to previous example. Pruned clusters using Formula \ref{formula:qualityClustering} are shadowed. Summary table gives information in relation to selected thresholds and colour gradients.}
\label{fig:cse-cic-2018-friday-paper}
\end{figure}
\end{landscape}

The tree depicted in Figure \ref{fig:cse-cic-2018-thursday-paper} offers us some insights on the process followed in computing it.

The basic ideas employed by the algorithm are practical and to the point. The fact that only a handful of the circa 80 features are used greatly contributes to the simplicity of the solution found. 

This can be partly because the generated traffic for the day under study, from the CSE-CIC-IDS 2018-thursday-15 dataset, was not too varied. Mainly composed of denial of service attacks \cite{ids-2018, ids-2018-web} (DoS-GoldenEye and Dos-Slowloris), it did not include brute force attacks, infiltration, nor SQL injections, for example. That said, these two attacks are very different, despite belonging to the same DoS family.

Both are HTTP-based and have in common to be relatively subtle, very targeted, and not relying on the generation of large volumes of traffic, as many other (D)DoS attacks do. This less sophisticated type of DoS is generally called flood-based or volumetric and is generally much easier to detect but harder to stop. 

This makes the classification of our two attacks under scrutiny much harder, as one can not rely on significant bandwidth changes at the start/end of the attack, nor on any side-effects that these high volumes of traffic may cause. In these two attacks, the forward direction refers to when the attacker sends traffic to the victim.

The main features used by the algorithm to build the tree are:
\begin{itemize}
    \item \emph{Tot Fwd Pkts} refers to the total number of packets in the forward direction, that is, from source to destination.  
    \item \emph{Active Mean} is the mean time a flow was active before becoming idle. 
    \item \emph{Bwd Pkt Len Std} relates to the standard deviation of the packet size in the backward direction. 
    \item \emph{Idle Std} is the standard deviation of the time a flow is idle before becoming active. 
    \item \emph{Flow Duration} is the duration of the flow.
    \item \emph{Fwd Pkts/s} is the number of forward packets per second.
    \item \emph{Flow IAT Std} refers to the standard deviation time between two flows.
    \item \emph{Active Min} describes the minimum time a flow is active before becoming idle.
    \item \emph{Bwd Header Len} is the total bytes used for headers in the backward direction.
\end{itemize}

We can observe how, in general, the tree tends to find increasing traffic volumes more indicative of anomalous or malign behaviour. For example, the vast majority of attack classifications lay to the right of the \emph{Tot Fwd Pkts} root node, indicating more exchanged packets. The algorithm combines both the volume of data exchanged in a connection and the time associated with it. This is what an expert would do, as both dimensions are highly critical in network attacks, notably those inducing denial of service (DoS). Again, certain types of DoS attacks may leverage the possibility of opening connections for very long periods of time, sending minimal data to keep them from turning idle, thus depleting server resources. The use of the \emph{Active Mean} feature captures this strategy. As seen in the tree, lower values of \emph{Active Mean} lead to a benign classification. Another choice by the algorithm is the use of the \emph{Bwd Pkt Len Std} feature. The algorithm does not care about the average value, just the standard deviation. When attacked frequently, the length of the packets returned is very similar, thus leading to a much smaller variance. This smart observation can probably be of general use in many different attack scenarios.

Somewhat related to this insight is the use of the \emph{Idle Std} feature. Again the algorithm focuses on the standard deviation instead of other first-order statistics. Although this feature is less valuable and is used deeper into the tree, it is employed twice in the pruned result. In this particular case, on both occasions, higher values lead to a more malign classification. On benign traffic, flows do not usually stay idle for a long time as this is wasteful, and both its average time and its standard deviation should be small. However, during DoS attacks (particularly Slowloris), these times tend to become much larger. The coexistence of benign traffic with Slowloris attacks makes these values wildly fluctuate between normal and atypical, thus increasing its standard deviation. Another interesting finding by the algorithm is the use of the \emph{Fwd Pkts/s} feature to detect a higher volume of traffic directed towards the victim, consistent with DoS attacks.

We continue with another example from the CSE-CIC-IDS 2018 dataset that, according to the documentation \cite{ids-2018-web}, corresponds to Friday the 16th of February 2018 when there were primarily two types of attacks over the network under study. These were named DoS-SlowHTTPTest and DoS-Hulk. As their name strongly suggests, both are again of the Denial of Service (DoS) type.

In the tree depicted in Figure \ref{fig:cse-cic-2018-friday-paper}, most of the malign traffic is in the left subtree, while the right subtree is further divided neatly into a subtree containing most of the more suspicious activity again to the left and primarily benign traffic to the right.

At the root of the tree, the value of the feature \emph{Flow IAT Mean} is interrogated. This refers to the average time between two flows. If this average time is too small, implying a burst-like traffic flow, then the chances are that this is malign. This is reasonable and in line with the HULK attack, where many short HTTP requests per second can increase the load of the web server. Following along the right subtree, the next feature considered is \emph{Fwd Pkts/s} which captures the number of packets per second in the flow in the forward direction, thus generated by the attacker. Higher throughputs correspond to a higher likelihood of attack, as expected. Same with \emph{Flow Duration}.

Moving to the right and leaving the vast majority of malicious traffic, the first attribute to decide here is \emph{Fwd Pkt Len Std}. This refers to the standard deviation of the packet length sent by the potential attacker. It is really insightful to note that when an attacker sends the same type of packet with minimal differences, this can be noticed very effectively by inspecting this feature. At least three of the four types of attack possible with SlowHTTPTest (see Appendix \ref{appendix:qualitative_assessment_trees}) and, to a certain extend also HULK, fit this description. It follows from this insight that lower values of this standard deviation led to somewhat more suspicious classifications, shown in the tree in the centre, in the subtree to the right. On the other hand, higher values of this standard deviation more closely correspond to normal, benign traffic, and as such most of the traffic represented by the subtree to the right is classified as innocuous. The last interesting element of this tree is the subtree below the \emph{Fwd IAT Mean}, which refers to the mean time between two packets sent in the forward (attacker to the web server) direction. This tends to be higher in attacks such as slow read from SlowHTTPTest. Though we address the node with a density of 42.08\%, the same analysis applies to a similar subtree with a density of 32.96\%.

One last observation is the use of the attribute \emph{Bwd Header Len}, which is related to the total bytes used for headers in the backward (mostly victim to the attacker) direction. Only high values in this feature are classified as malign. The algorithm grasps the strange behaviour in the packet headers when being under an attack by some of the SlowHTTPTest tools. This shows that the algorithm can detect, recognise, and classify the type of attack observed, as this abnormal header manipulation is characteristic of three out of the four SlowHTTPTest attack types. This approach allows the clustering of different attack types to reveal the most characteristic features of each attack studied. This could help an expert identify the main elements of an attack and, thanks to the explainability of the trees, quickly turn these into rules that will detect specific attacks.

Lastly, briefly delving into the results of the ISCXIDS 2012 dataset$^7$, the feature at the root of the tree is \emph{totalSourceBytes}. This measures the total amount of data sent by the communication source in bytes.

This concept can be interpreted as an insightful way of detecting attacks in the dataset. Many attack attempts, particularly those belonging to the (D)DoS family, need to send lots of data to their targets. This feature divides the tree into two halves. The one to the left corresponds to communications where the sender submitted an amount of data at or below a threshold (in this case, 5900.5 bytes). That puts most of the benign traffic to the left tree leaves. On the other hand, the majority of the right subtree paths result in attacks.

Previous discussion on \emph{totalSourceBytes} also applies, with some minor caveats, to other related attributes such as \emph{totalDestinationBytes}, \emph{totalSourcePackets} or \emph{totalDestinationPackets}. These are also used extensively through the decision tree. The right side of the tree only accounts for around 4.74\% of the total traffic seen, most of it attack-related, characterised as byte-dense.

On the subtree to the left, we can observe that the next deciding feature is \emph{appName}. This attribute contains the name of the level 7 application targeted in the communication. Many attacks are very application-specific. If we can discern which applications and services are most frequently targeted, this can be of great help in filtering out attacks. These applications or services can be an attractive target for an attacker on a temporal basis (e.g., due to a recent 0-day or new exploit) or on a more permanent one, like DNS, SSH, SMTP, HTTP/S, FTP, etc. A combination of both strategies can be seen in the particular set picked by the algorithm, including \emph{\{Unknown\_UDP, SecureWeb, SAP, and FTP\}}. Deeper in the subtree to the right, the algorithm uses a similar approach based on the values of the same feature.

This repeated filtering by \emph{appName} offers excellent results from a classification point of view of the attacks. This is a strategy that humans use and reflect on some widely deployed defence mechanisms (e.g., firewall rules).

\section{Conclusions}\label{sec:conclusions}
XAI goes well beyond social or legal issues. Explanation algorithms are necessary to provide transparency and to justify model predictions. This work exemplifies the ``black box" behaviour problem focusing on the anomaly detection algorithm ADMNC. We propose an extended framework, EADMNC, which comprises a new layer to avoid these shortcomings and understand trained models. XAI is a recent and growing field of study, and it can be instrumental in providing guarantees of compliance with, among others, European Union regulations.

\begin{itemize}
    \item Our algorithm provides pre hoc and post hoc explanations for anomalies detected on continuous and categorical variables in a unified way.
    \item Regression tree cluster descriptions allow us to examine differences between normal and anomalous data.
    \item Further study about the validity and usefulness of our approach is provided, reasoning about selected variables by the algorithm in the network intrusion detection domain.
    \item Text-based post hoc explanations offer justifications for positive detections using both categorical and continuous data. We leveraged the formulation of ADMNC to design a Rule-based system to obtain explanations for each anomaly detected.
    \item In the pruned regression tree, the paths to the clusters within the anomalous estimator range give us the information to describe and explain anomalous conditions.
    \item The implementation using the Apache Spark framework maintains the scalability of the original ADMNC, parallelizing stages during the training of both models.
\end{itemize}

One possible future research line is to improve explanations by introducing a preliminary dimensionality reduction step. This idea makes sense because high dimensional data generally present redundant and irrelevant variables that can produce bias and generalisation errors. Techniques such as PCA (Principal Component Analysis) \cite{PCA_Anomaly} can be used to transform input variables into a non-correlated reduced set of them. The objective is to provide an explanation using data variance information. Another option is FS (Feature Selection) \cite{FeaturesSelection_online, FeatureSelection_review_state_of_art}, which extracts the most relevant variables of the input vectors, simplifying prediction models and thus making explanations easier. This FS approach can also significantly reduce computation time on high dimensional data. Additionally, filtering redundant data can favour generalisation capabilities and help to avoid over-fitting. 
Finally, another novel research line could be to improve the two approaches for text-based explanations. One possibility could be adding more rules. This approach will undoubtedly cover more situations. Offering higher-level explanations for end-users is a medium-term objective, also guaranteeing the robustness and generalisation capabilities of our method.

\section*{Acknowledgements}

This research has been financially supported in part by the Spanish Government research project PID2019-109238GB-C22, by the Xunta de Galicia under grant ED431G 2019/01, and by European Union ERDF Funds.  Special recognition goes to the Spanish Ministerio de Educación for the predoctoral FPU funds, grant number FPU19/01457. We would also like to thank CESGA (Centro de Supercomputación de Galicia) for giving us access to their computing resources.

\ifCLASSOPTIONcaptionsoff
  \newpage
\fi



\bibliographystyle{unsrt}
\bibliography{bibliography}

\begin{thebibliography}{10}

\bibitem{hawkingsBook1}
D.~M. Hawkings.
\newblock {\em Identification of outliers}, volume~11.
\newblock Springer, 1980.

\bibitem{Hodge04}
V.~J. Hodge and J.~Austin.
\newblock A survey of outlier detection methodologies.
\newblock {\em Artificial Intelligence Review}, 22:85--126, 2004.

\bibitem{kumarBook1}
V.~{Kumar}.
\newblock Parallel and distributed computing for cybersecurity.
\newblock {\em IEEE Distributed Systems Online}, 6(10), 2005.

\bibitem{vigilanciaPaper}
A.~A. {Sodemann}, M.~P. {Ross}, and B.~J. {Borghetti}.
\newblock {A Review of Anomaly Detection in Automated Surveillance}.
\newblock {\em IEEE Transactions on Systems, Man, and Cybernetics, Part C
  (Applications and Reviews)}, 42(6):1257--1272, Nov 2012.

\bibitem{Francos13}
D.~Fernández-Francos, D.~Martínez-Rego, O.~Fontebla-Romero, and
  A.~Alonso-Betanzos.
\newblock Automatic bearing fault diagnosis based on one-class $\nu$-svm.
\newblock {\em Computers \& Industrial Engineering}, 64:357--365, 2013.

\bibitem{explicacion_paper_XAI}
A.~{Adadi} and M.~{Berrada}.
\newblock {Peeking Inside the Black-Box: A Survey on Explainable Artificial
  Intelligence} ({XAI}).
\newblock {\em IEEE Access}, 6:52138--52160, 2018.

\bibitem{Samek19}
W.~Samek and K.~R. Müller.
\newblock {\em Towards Explainable Artificial Intelligence.}, volume 11700 of
  {\em Lecture Notes in Computer Science}, pages 5--22.
\newblock Springer, 2019.

\bibitem{darpa_xai}
D.~Gunning.
\newblock {Explainable Artificial Intelligence} ({XAI}), mar 2018.
\newblock
  \href{https://www.darpa.mil/program/explainable-artificial-intelligence}{darpa.mil
  }{[Online; posted March-2018]}.

\bibitem{state-of-art_XAI}
S.~T. Mueller, R.~R. Hoffman, W.~J. Clancey, A.~Emrey, and G.~Klein.
\newblock {Explanation in Human-AI Systems}: {A} {Literature Meta-Review,
  Synopsis of Key Ideas and Publications, and Bibliography for Explainable}
  {AI}.
\newblock {\em CoRR}, abs/1902.01876, 2019.

\bibitem{HLEGAI19}
High Level Expert~Group on~Artificial~Intelligence.
\newblock {Ethics Guidelines on Trustworthy Artificial Intelligence}, 2019.
\newblock
  \href{https://ec.europa.eu/digital-single-market/en/news/ethics-guidelines-trustworthy-ai}{[Online;
  posted April-2019}.

\bibitem{reglamento_general_proteccion_datos}
A.~Calder.
\newblock {\em EU GDPR: A Pocket Guide}.
\newblock IT Governance Publishing, 2016.

\bibitem{cita_reglamento_eu}
Official~Journal of~the European~Union.
\newblock Regulation ({EU}) 2016/679 of the european parliament and of the
  council of 27 april 2016 on the protection of natural persons with regard to
  the processing of personal data, apr 2016.
\newblock
  \href{https://eur-lex.europa.eu/legal-content/EN/TXT/PDF/?uri=CELEX:32016R0679&from=EN}{eur-lex.europa.eu
  }{[Online; posted 27-April-2016]}.

\bibitem{referencia_figuras_xai_eml}
A.~Calvo.
\newblock {We are ready for Machine Learning Explainability?}, apr 2019.
\newblock
  \href{https://towardsdatascience.com/we-are-ready-to-ml-explainability-2e7960cb950d}{towardsdatascience.com
  }{[Online; posted 8-April-2019]}.

\bibitem{Samek-book}
W.~Samek, G.~Montavo, A.~Vedaldi, L.~K. Hansen, and K.~R. Müller.
\newblock {\em Explainable AI: Interpreting, Explaining and Visualizing Deep
  Learning}.
\newblock Springer, Lecture Notes in Artificial Intelligence 11700, 2019.

\bibitem{explanation_interpretation_definitions}
R.~Marcinkevics and J.~E. Vogt.
\newblock {Interpretability and Explainability}: {A} {Machine Learning Zoo
  Mini-tour}.
\newblock {\em CoRR}, abs/2012.01805, 2020.

\bibitem{rudin_explainability}
C.~Rudin.
\newblock Stop explaining black box machine learning models for high stakes
  decisions and use interpretable models instead.
\newblock {\em Nature Machine Intelligence}, 1(5):206--215, 2019.

\bibitem{admnc}
C.~Eiras-Franco, D.~Martínez-Rego, B.~Guijarro-Berdiñas, A.~Alonso-Betanzos,
  and A.~Bahamonde.
\newblock Large scale anomaly detection in mixed numerical and categorical
  input spaces.
\newblock {\em Information Sciences}, 487:115 -- 127, 2019.

\bibitem{review_general_explainable_surrogate}
{A. Ravi, X. Yu, I. Santelices, F. Karray and B. Fridan}.
\newblock {General Frameworks for Anomaly Detection Explainability: Comparative
  Study}.
\newblock In {\em 2021 IEEE International Conference on Autonomous Systems
  (ICAS)}, pages 1--5, 2021.

\bibitem{lime}
{M. T. Ribeiro, S. Singh and C. Guestrin}.
\newblock "{Why} {Should} {I} {Trust You?": Explaining the Predictions of Any
  Classifier}.
\newblock In {\em Proceedings of the 22nd {ACM} {SIGKDD} International
  Conference on Knowledge Discovery and Data Mining, San Francisco, CA, USA,
  August 13-17, 2016}, pages 1135--1144, 2016.

\bibitem{lime_use_case_1_one_class_ae}
{T. Wu and Y. Wang}.
\newblock {Locally Interpretable One-Class Anomaly Detection for Credit Card
  Fraud Detection}.
\newblock {\em CoRR}, abs/2108.02501:1--6, 2021.

\bibitem{lime_shap_both_use_case}
{I. Psychoula, A. Gutmann, P. Mainali, S. H. Lee, P. Dunphy and F. Petitcolas}.
\newblock {Explainable Machine Learning for Fraud Detection}.
\newblock {\em Computer}, 54(10):49--59, 2021.

\bibitem{Shap}
{S. M. Lundberg and S. Lee}.
\newblock {A Unified Approach to Interpreting Model Predictions}.
\newblock In I.~Guyon, U.~V. Luxburg, S.~Bengio, H.~Wallach, R.~Fergus,
  S.~Vishwanathan, and R.~Garnett, editors, {\em Advances in Neural Information
  Processing Systems 30}, pages 4765--4774. Curran Associates, Inc., 2017.

\bibitem{tree_shap}
{S. M. Lundberg, G. Erion, H. Chen, A. DeGrave, J. M. Prutkin, B. Nair, R.
  Katz, J. Himmelfarb, N. Bansal and S. Lee}.
\newblock {From local explanations to global understanding with explainable AI
  for trees}.
\newblock {\em Nature Machine Intelligence}, 2(1):56--67, 2020.

\bibitem{gpu_tree_shap}
{R. Mitchell, E. Frank and G. Holmes}.
\newblock {GPUTreeShap: Fast Parallel Tree Interpretability}.
\newblock {\em CoRR}, abs/2010.13972(2):1--10, 2020.

\bibitem{shap_use_1}
{O. Serradilla, E. Zugasti, J. R. de Okariz, J. Rodriguez and U. Zurutuza}.
\newblock {Adaptable and explainable predictive maintenance: semi-supervised
  deep learning for anomaly detection and diagnosis in press machine data}.
\newblock {\em Applied Sciences (Switzerland)}, 11(16), 2021.

\bibitem{shap_diffi_use_case}
{L. C. Brito, G. A. Susto, J. N. Brito and M.A.V. Duarte}.
\newblock {An explainable artificial intelligence approach for unsupervised
  fault detection and diagnosis in rotating machinery}.
\newblock {\em Mechanical Systems and Signal Processing}, 163(March
  2021):108105, 2022.

\bibitem{shap_use_case_2}
{S. Park, J. Moon and E. Hwang}.
\newblock {Explainable Anomaly Detection for District Heating Based on Shapley
  Additive Explanations}.
\newblock {\em IEEE International Conference on Data Mining Workshops, ICDMW},
  2020-Novem:762--765, 2020.

\bibitem{shap_use_case_3}
{L. Antwarg, R. M. Miller, B. Shapira and L. Rokach}.
\newblock {Explaining Anomalies Detected by Autoencoders Using SHAP}.
\newblock {\em CoRR}, abs/1903.02407:1--37, 2019.

\bibitem{shap_use_case_4}
{J. Jakubowski, P. Stanisz, S. Bobek and G. J. Nalepa}.
\newblock Explainable anomaly detection for hot-rolling industrial process.
\newblock In {\em 2021 IEEE 8th International Conference on Data Science and
  Advanced Analytics (DSAA)}, pages 1--10, 2021.

\bibitem{lrp_original}
{S. Bach, A. Binder, G. Montavon, F. Klauschen, K. R. M{\"{u}}ller, and W.
  Samek}.
\newblock {On pixel-wise explanations for non-linear classifier decisions by
  layer-wise relevance propagation}.
\newblock {\em PLoS ONE}, 10(7):1--46, 2015.

\bibitem{lrp_overview}
{G. Montavon, A. Binder, S. Lapuschkin, W. Samek and K. R. M{\"u}ller}.
\newblock {\em Layer-Wise Relevance Propagation: An Overview}, pages 193--209.
\newblock Springer International Publishing, Cham, 2019.

\bibitem{lrp_extension}
{A. Binder, G. Montavon, S. Lapuschkin, K. R. M{\"{u}}ller and W. Samek}.
\newblock {Layer-wise relevance propagation for neural networks with local
  renormalization layers}.
\newblock {\em Lecture Notes in Computer Science (including subseries Lecture
  Notes in Artificial Intelligence and Lecture Notes in Bioinformatics)}, 9887
  LNCS:63--71, 2016.

\bibitem{lrp_use_case_1}
{M. Kohlbrenner, A. Bauer, S. Nakajima, A. Binder, W. Samek, and S.
  Lapuschkin}.
\newblock {Towards Best Practice in Explaining Neural Network Decisions with
  LRP}.
\newblock {\em Proceedings of the International Joint Conference on Neural
  Networks}, 2020.

\bibitem{lrp_use_case_2}
{A. Patil, A. Wadekar, T. Gupta, R. Vijan and F. Kazi}.
\newblock {Explainable LSTM Model for Anomaly Detection in HDFS Log File using
  Layerwise Relevance Propagation}.
\newblock {\em 2019 IEEE Bombay Section Signature Conference, IBSSC 2019},
  2019Januar, 2019.

\bibitem{exad_framework}
{B. Rad, F. Song, V. Jacob and Y. Diao}.
\newblock {Explainable anomaly detection on high-dimensional time series data}.
\newblock {\em DEBS 2021 - Proceedings of the 15th ACM International Conference
  on Distributed and Event-Based Systems}, 1:142--147, 2021.

\bibitem{deep_vs_classic_shallow_AD}
{L. Ruff, J. R. Kauffmann, R. A. Vandermeulen, G. Montavon, W. Samek, M. Kloft,
  T. G. Dietterich, and K. R. Muller}.
\newblock {A Unifying Review of Deep and Shallow Anomaly Detection}.
\newblock {\em Proceedings of the IEEE}, 109(5):756--795, 2021.

\bibitem{quirkExample}
T.~{Mokoena}, O.~{Lebogo}, A.~{Dlaba}, and V.~{Marivate}.
\newblock Bringing sequential feature explanations to life.
\newblock In {\em 2017 IEEE AFRICON}, pages 59--64, Sep. 2017.

\bibitem{LODI_explain}
{X. H. Dang, B. Micenkov{\'a}, I. Assent and R. T. Ng}.
\newblock Local outlier detection with interpretation.
\newblock In H.~Blockeel, K.~Kersting, S.~Nijssen, and F.~{\v{Z}}elezn{\'y},
  editors, {\em Machine Learning and Knowledge Discovery in Databases}, pages
  304--320, Berlin, Heidelberg, 2013. Springer Berlin Heidelberg.

\bibitem{LOGP_explain}
{X. H. {Dang}, I. \mbox{Assent}, R. T. {Ng}, A. {Zimek} and E.
  \mbox{Schubert}}.
\newblock Discriminative features for identifying and interpreting outliers.
\newblock In {\em 2014 IEEE 30th International Conference on Data Engineering},
  pages 88--99, March 2014.

\bibitem{explain_anomaly_general_methods}
{B. \mbox{Micenková}, R. T. {Ng}, X. H. {Dang} and I. \mbox{Assent}}.
\newblock {Explaining Outliers by Subspace Separability}.
\newblock In {\em 2013 IEEE 13th International Conference on Data Mining},
  pages 518--527, Dec 2013.

\bibitem{one_class_svm_rule_extractor}
{A. Barbado, O. Corcho and R. Benjamins}.
\newblock {Rule extraction in unsupervised anomaly detection for model
  explainability: Application to OneClass SVM}.
\newblock {\em Expert Systems with Applications}, 189(October 2021):116100,
  2022.

\bibitem{diffi_iforest_interpretation}
{M. Carletti, C. Masiero, A. Beghi and G. A. Susto}.
\newblock {Explainable machine learning in industry 4.0: Evaluating feature
  importance in anomaly detection to enable root cause analysis}.
\newblock {\em Conference Proceedings - IEEE International Conference on
  Systems, Man and Cybernetics}, 2019-Octob:21--26, 2019.

\bibitem{iforest_explanator_2}
{N. S. Kartha, C. Gautrais, V. Vercruyssen and K. U. Leuven}.
\newblock {Why Are You Weird? Infusing Interpretability in Isolation Forest for
  Anomaly Detection}.
\newblock {\em CoRR}, abs/2112.06858:1--8, 2021.

\bibitem{iforest_anomaly_detection}
{F. T. Liu, K. M. Ting and Z.H. Zhou}.
\newblock {Isolation-Based Anomaly Detection}.
\newblock {\em ACM Trans. Knowl. Discov. Data}, 6(1), mar 2012.

\bibitem{lstm_gradient_boosted_decision_trees}
{J. Chatterjee and N. Dethlefs}.
\newblock {Deep learning with knowledge transfer for explainable anomaly
  prediction in wind turbines}.
\newblock {\em Wind Energy}, 23(8):1693--1710, 2020.

\bibitem{deep_alternatives_anomaly_detection}
{M. Gribbestad, M. U. Hassan, I. A. Hameed and K. Sundli}.
\newblock {Health monitoring of air compressors using reconstruction-based deep
  learning for anomaly detection with increased transparency}.
\newblock {\em Entropy}, 23(1):1--27, 2021.

\bibitem{deep_explainable_anomaly_detection}
{G. Pang and C. Aggarwal}.
\newblock {\em {Toward Explainable Deep Anomaly Detection}}, volume~1.
\newblock Association for Computing Machinery, 2021.

\bibitem{fcdd_heatmap_anomaly_explain}
{P. Liznerski, L. Ruff, R. A. Vandermeulen, B. J. Franks, M. Kloft, and K. R.
  M{\"{u}}ller}.
\newblock {Explainable Deep One-Class Classification}.
\newblock {\em CoRR}, abs/2007.01760:1--25, jul 2020.

\bibitem{deep_anomaly_heatmap_extraction}
{S. Kitamura and Y. Nonaka}.
\newblock {\em {Explainable Anomaly Detection via Feature-Based Localization}},
  volume 11731 LNCS.
\newblock Springer International Publishing, 2019.

\bibitem{deep_cae_explain}
{R. Assaf, I. Giurgiu, J. Pfefferle, S. Monney, H. Pozidis and A.Schumann}.
\newblock {An anomaly detection and explainability framework using
  convolutional autoencoders for data storage systems}.
\newblock {\em IJCAI International Joint Conference on Artificial
  Intelligence}, 2021-Janua:5228--5230, 2020.

\bibitem{gee_vae_network_intrusion}
{Q. P. Nguyen, K. W. Lim, D. M. Divakaran, K. H. Low and M. C. Chan}.
\newblock {GEE: A Gradient-based Explainable Variational Autoencoder for
  Network Anomaly Detection}.
\newblock {\em 2019 IEEE Conference on Communications and Network Security, CNS
  2019}, pages 91--99, 2019.

\bibitem{devnet_anomaly_scores_explanation}
{G. Pang, C. Ding, C. Shen and A. van den Hengel}.
\newblock {Explainable Deep Few-shot Anomaly Detection with Deviation
  Networks}.
\newblock {\em CoRR}, abs/2108.00462:1--16, 2021.

\bibitem{DAART_explain}
A.~Smith-Renner, R.~Rua, and M.~Colony.
\newblock {Towards an Explainable Threat Detection Tool}.
\newblock In {\em IUI Workshops}, 2019.

\bibitem{book_CART_explained}
{X. Wu, V. Kumar, J. Ross Quinlan, J. Ghosh, Q. Yang, H. Motoda, G. J.
  McLachlan, A. Ng, B. Liu, P. S. Yu, Z. H. Zhou, M. Steinbach, D. J. Hand and
  D. Steinberg}.
\newblock Top 10 algorithms in data mining.
\newblock {\em Knowledge and Information Systems}, 14(1):1--37, Jan 2008.

\bibitem{EMgaussians}
I.~F. {Iatan}.
\newblock The expectation-maximization algorithm: Gaussian case.
\newblock In {\em 2010 International Conference on Networking and Information
  Technology}, pages 590--593, June 2010.

\bibitem{oneHotCoding}
M.~{Cassel} and F.~{Lima}.
\newblock {Evaluating one-hot encoding finite state machines for SEU
  reliability in SRAM-based FPGAs}.
\newblock In {\em 12th IEEE International On-Line Testing Symposium
  (IOLTS'06)}, pages 1--6, July 2006.

\bibitem{estocasticGradientDescent}
L.~Bottou and O.~Bousquet.
\newblock {The Tradeoffs of Large Scale Learning}.
\newblock In J.~C. Platt, D.~Koller, Y.~Singer, and S.~T. Roweis, editors, {\em
  Advances in Neural Information Processing Systems 20}, pages 161--168. Curran
  Associates, Inc., 2008.

\bibitem{mini_batch_step_gradient_desc}
A.~Cotter, O.~Shamir, N.~Srebro, and K.~Sridharan.
\newblock {Better Mini-Batch Algorithms via Accelerated Gradient Methods}.
\newblock In J.~Shawe-Taylor, R.~S. Zemel, P.~L. Bartlett, F.~Pereira, and
  K.~Q. Weinberger, editors, {\em Advances in Neural Information Processing
  Systems 24}, pages 1647--1655. Curran Associates, Inc., 2011.

\bibitem{darpa_xai_decision_trees}
D.~Gunning.
\newblock {DARPA’s Explainable Artificial Intelligence ({XAI}) Program}.
\newblock In {\em Proceedings of the 24th International Conference on
  Intelligent User Interfaces}, IUI ’19, page~2, New York, NY, USA, 2019.
  Association for Computing Machinery.

\bibitem{decision_trees_explanation_study}
S.~M. Lundberg, G.~G. Erion, H.~Chen, A.~J. DeGrave, J.~M Prutkin, B.~G. Nair,
  R.~Katz, J.~Himmelfarb, N.~Bansal, and S.~I. Lee.
\newblock {Explainable {AI} for Trees: From Local Explanations to Global
  Understanding}.
\newblock {\em ArXiv}, abs/1905.04610, 2019.

\bibitem{dyadic_data_explain}
C.~Eiras-Franco, B.~Guijarro-Berdiñas, A.~Alonso-Betanzos, and A.~Bahamonde.
\newblock A scalable decision-tree-based method to explain interactions in
  dyadic data.
\newblock {\em Decision Support Systems}, page 113141, 2019.

\bibitem{BBDD_datasets}
D.~Dua and C.~Graff.
\newblock {UCI} {Machine Learning Repository}.
\newblock University of California, Irvine, School of Information and Computer
  Sciences, 2017.

\bibitem{kdd-cup-99}
S.~{Hettich} and S.~{Bay}.
\newblock {KDD} cup 1999 data.
\newblock The UCI KDD Archive, Irvine, CA: University of California, Department
  of Information and Computer Science., 1999.

\bibitem{ids-2012}
A.~Shiravi, H.~Shiravi, M.~Tavallaee, and A.~A. Ghorbani.
\newblock Toward developing a systematic approach to generate benchmark
  datasets for intrusion detection.
\newblock {\em Computers \& Security}, 31(3):357 -- 374, 2012.

\bibitem{ids-2018}
I.~Sharafaldin, A.~H. Lashkari, and A.~A. Ghorbani.
\newblock {Toward Generating a New Intrusion Detection Dataset and Intrusion
  Traffic Characterization}.
\newblock In {\em ICISSP}, 2018.

\bibitem{ids-2018-web}
Communications Security Establishment (CSE)~\& the Canadian Institute~for
  Cybersecurity~(CIC).
\newblock {CSE-CIC-IDS-2018} on {AWS}, mar 2018.
\newblock \href{https://www.unb.ca/cic/datasets/ids-2018.html}{unb.ca
  }{[Online; posted March-2018]}.

\bibitem{auroc_Melo2013}
F.~Melo.
\newblock {\em Area under the ROC Curve}, pages 38--39.
\newblock Springer New York, New York, NY, 2013.

\bibitem{PCA_Anomaly}
L.~Huang, X.~Nguyen, M.~Garofalakis, M.~I. Jordan, A.~Joseph, and N.~Taft.
\newblock {In-Network PCA and Anomaly Detection}.
\newblock In {\em Proceedings of the 19th International Conference on Neural
  Information Processing Systems}, NIPS’06, page 617–624, Cambridge, MA,
  USA, 2006. MIT Press.

\bibitem{FeaturesSelection_online}
S.~G. {Devi} and M.~{Sabrigiriraj}.
\newblock {Feature Selection, Online Feature Selection Techniques for Big Data
  Classification: A Review}.
\newblock In {\em 2018 International Conference on Current Trends towards
  Converging Technologies (ICCTCT)}, pages 1--9, March 2018.

\bibitem{FeatureSelection_review_state_of_art}
S.~{Visalakshi} and V.~{Radha}.
\newblock A literature review of feature selection techniques and applications:
  Review of feature selection in data mining.
\newblock In {\em 2014 IEEE International Conference on Computational
  Intelligence and Computing Research}, pages 1--6, Dec 2014.

\end{thebibliography}


%










\appendices

\section{Detailed analysis of the combined post hoc explanation}
\label{appendix:detailed_post_hoc_explanation}

This text starts with a description of the detected anomaly, mainly giving details about continuous and categorical vectors. Firstly, it summarises details of the detection correlated with continuous data (1). That is, the predicted Gaussian corresponding to the GMM model \cite{admnc}, the continuous estimator obtained from the factorisation of the pdf given in Formula \ref{formula:general_pdf_factorization} and which rule is fired using the Rule-based classification System. Secondly, about the correlation with categorical data (2), the logistic estimator obtained from the factorisation of the pdf is given again, this time focusing on the already explained decomposition of Formula \ref{form:general_lr}. In this way, the average estimator of the categorical terms is presented, as well as the subset of categorical terms fulfilling hyper-parameter constraints ($T_{filter}$). We made the correspondence with the original categorical values providing, when available, the correlation with continuous data based on the conditional probability.

\begin{lstlisting}[language=bash, keywords={smooth}, basicstyle=\small\ttfamily, mathescape=true]
Detected anomaly N(6):
** Continuous vector details (1):
$\rightarrow$ Predicted Gaussian (class): 0
$\rightarrow$ Continuous anomalous estimator: -7924.463
$\rightarrow$ Rule-based explanation (1):
$\longrightarrow$ First rule is fired -- It is an anomaly since continuous sample is clearly separated from learned groups.
$\longrightarrow$ Continuous pattern information: probability of belonging to each of the Gaussians (classes from 0 to 1):
$\longrightarrow$ pdf(class=0) = -7924.463
$\longrightarrow$ pdf(class=1) = -7924.463
** Categorical vector details (2):
$\rightarrow$ Logistic estimator: -14.469
$\rightarrow$ Average categorical estimator: 0.543
$\rightarrow$ Number of categorical estimators detected below anomalous threshold: 2
$\longrightarrow$ [1/2] It is an anomaly since categorical feature "Existence of ragged T wave" has a value of 0.0
$\longrightarrow$ [1/2] Categorical estimator value: 0.322
$\longrightarrow$ [2/2] It is an anomaly since categorical feature "Existence of ragged P wave" has a value of 0.0
$\longrightarrow$ [2/2] Categorical estimator value: 0.357
$\longrightarrow$ [2/2] Involved continuous feature "channel_V5_amplitude_Rwave" with value -0.525.
\end{lstlisting}


\section{}
\label{appendix:table_q_lambda}
\begin{table}[!htb]
\caption{$\mathcal{Q}$ (quality) values following the format $P$($F$). $P$ stands for tree after pruning (Pruned) and $F$ before pruning (Full). $\boldsymbol{\lambda}$ values used in the pruning algorithm are also provided.}
\label{table:metrics_q_lambda}
\centering
\begin{tabular}{|l|c|c|}
\hline
\textbf{Dataset} & \multicolumn{2}{|c|}{\textbf{Metrics}} \\ \hline
\textbf{Name} & $\boldsymbol{\lambda}$ & \boldsymbol{$\mathcal{Q}$} \\ \hline
Arrhythmia & $10^{-4}$ & $-0.013(-0.021)$ \\ \hline
German Credit & $10^{-4}$ & $-0.004(-0.014)$ \\ \hline
Ab. 1 & $10^{-5}$ & $-0.007(-0.008)$ \\ \hline
Ab. 9 & $10^{-5}$ & $-0.005(-0.006)$ \\ \hline
Ab. 11 & $10^{-5}$ & $-0.006(-0.007)$ \\ \hline
IDS-2012 & $10^{-8}$ & $-7.56\times10^{-6}(-8.05\times10^{-6})$ \\ \hline
IDS-2018-F & $10^{-4}$ & $-0.006(-0.018)$ \\ \hline
IDS-2018-T & $10^{-7}$ & $-1.64\times10^{-5}(-2.56\times10^{-5})$ \\ \hline
IDS-2018-W & $10^{-5}$ & $-0.003(-0.004)$ \\ \hline
KDD-FULL & $10^{-5}$ & $-0.011(-0.012)$ \\ \hline
KDD-SMTP & $10^{-8}$ & $-8.78\times10^{-7}(-2.12\times10^{-6})$ \\ \hline
KDD-HTTP & $10^{-8}$ & $-1.77\times10^{-7}(-1.67\times10^{-6})$ \\ \hline
KDD-10 & $10^{-8}$ & $-4.73\times10^{-6}(-1.77\times10^{-5})$ \\ \hline
Synth-1 NV & $10^{-6}$ & $-0.0002(-0.0003)$ \\ \hline
Synth-2 NV & $10^{-6}$ & $-0.0001(-0.0002)$ \\ \hline
Synth-4 NV & $10^{-6}$ & $-0.0002(-0.0002)$ \\ \hline
Synth-8 NV & $10^{-6}$ & $-0.0004(-0.0005)$  \\ \hline
Synth-16 NV & $10^{-6}$ & $-0.0003(-0.0003)$ \\ \hline
Synth-32 NV & $10^{-6}$ & $-0.0003(-0.0004)$ \\ \hline
\end{tabular}
\end{table}

\section{additional comments related to the trees}
\label{appendix:qualitative_assessment_trees}

\subsection{\textit{CSE-CIC-2018-thursday}}
Slowloris keeps as many open connections to the target web server as possible for as long as possible, so that new, legitimate users are affected. This is achieved by sending partial HTTP headers from time to time but never completing requests, so that the concurrent connection pool is filled and cannot accept new requests from legitimate users. This does not require large bandwidths or the generation of gigabytes of data.

GoldenEye$^8$ is a HTTP DoS Test Tool that exploits the ``HTTP Keep-Alive + NoCache" attack vector. This combination ensures that HTTP connections are kept open for the longest time possible, avoiding any cache speed-up by adding random useless characters to every request to look different for cache algorithms despite targeting the same resource. This similarly slows down or stops new connections from legitimate users without generating inordinate amounts of traffic.

\subsection{\textit{CSE-CIC-2018-friday}}
The SlowHTTPTest tool takes advantage of a feature in HTTP design, which waits until HTTP requests are complete before processing them. If a server receives incomplete HTTP requests or receives them at a plodding pace, the server allocates some resources to this request and does not free them until the request is finished. If the attacker submits many of these requests, the server can run out of resources and stop processing legitimate ones. This tool incorporates 4 different types of attack$^9$, all using the same principles but with slight variations in their execution and strategy. These are Slowloris (sending incomplete headers lacking the final CRLF), RUDeadYet (POST petitions with incomplete headers and a misleading Content-Length value), Apache killer (contradictory and misleading HTTP ranges in the header), and finally slow read (reasonable HTTP requests, but very delayed ACK).

HULK stands for ``Http Unbearable Load King". It will generate many uniquely crafted HTTP requests, each of which will slightly increase the load on the CPU of the web server until it crashes or fails to respond. It makes the requests dynamic and thus more challenging to detect with defensive signatures. For instance, HULK will rotate both User-Agent and Referer fields$^{10}$. This tool employs some basic evasion techniques based on randomising payloads that make its detection slightly more involved.


\section*{Footnotes}

\begin{enumerate}
    \item We generated texts based on tree transcriptions and also using basic text templates, considering variants and adapting them to each particular detection. This method is not related to the text generation field.
    \item Example from IDS-2018-W dataset, simplifying parts of the original text.
    \item Example from Arrhythmia dataset, simplifying parts of the original text.
    \item We decided to choose the same value for all the datasets for the sake of simplicity.
    \item Processed datasets are available at \url{https://udcgal-my.sharepoint.com/:f:/g/personal/inigo_lopezrioboo_botana_udc_es/ErgnMJsq_7FJs07Qkl8_vkoBzS_Pc995NCUmL_H-0ADdLw?e=2OjUvT}.
    \item Results are available at \url{https://www.dropbox.com/sh/m6lyn8zpss75sru/AADO_OFwzNwUTHD24vgJXhwma?dl=0}.
    \item Available at the previously referenced repository.
    \item Check \url{https://kb.mazebolt.com/knowledgebase/goldeneye-http-flood/} for a technical analysis.
    \item We encourage the reader to check \url{https://www.hackplayers.com/2016/06/ataques-dos-slow-http-mediante-SlowHTTPTest.html} and \url{https://github.com/shekyan/slowhttptest/wiki} for further information.
    \item \url{https://www.trustwave.com/en-us/resources/blogs/spiderlabs-blog/hulk-vs-thor-application-dos-smackdown/}.
\end{enumerate}

\end{document}